\documentclass[11pt]{article}

\usepackage[T1]{fontenc}
\usepackage{lmodern}
\usepackage[letterpaper,margin=1in]{geometry}
\usepackage{amsmath}
\usepackage{amssymb}
\usepackage{booktabs}
\usepackage{adjustbox}
\usepackage{float}
\usepackage{algorithm}
\usepackage{algpseudocode}
\usepackage{subcaption}
\usepackage{placeins}
\usepackage{graphicx}
\usepackage{microtype}
\usepackage[round]{natbib}
\usepackage[hidelinks]{hyperref}

\hypersetup{
  pdftitle={Generative Models as Adaptive Trust Regions for Discrete Bayesian Optimisation},
  pdfauthor={Michael Dodds, Erik Bodin, Sebastian W. Ober, Tom Diethe, Carl Henrik Ek, Henry Moss}
}

\title{Generative Models as Adaptive Trust Regions\\
for Discrete Bayesian Optimisation}
\author{Michael Dodds\textsuperscript{1},
Erik Bodin\textsuperscript{2},
Sebastian W.~Ober\textsuperscript{3},
Tom Diethe\textsuperscript{4},\\
Carl Henrik Ek\textsuperscript{1},
Henry Moss\textsuperscript{5}\\[1.5ex]
\normalsize \textsuperscript{1}University of Cambridge\quad
\textsuperscript{2}Monumo\quad
\textsuperscript{3}BigHat Biosciences\\
\normalsize \textsuperscript{4}AstraZeneca\quad
\textsuperscript{5}Lancaster University}
\date{}

\begin{document}

\maketitle

\begin{abstract}
Bayesian optimisation (BO) is a powerful framework for sample-efficient
optimisation of expensive black-box functions, but its application to
high-dimensional discrete scientific design remains challenging. Local trust-region BO
methods such as TuRBO address high dimensionality by restricting search to adaptively sized neighbourhoods.
However, a direct discrete analogue based on edit distance
can be poorly suited to structured discrete objects: neighbourhoods grow
combinatorially, and metric proximity does not capture the dependencies that
determine whether a design is plausible or useful. Motivated by the growing
availability of pretrained generative models, we introduce
GenTuRBO, which replaces the geometric trust region with an adaptively scaled
proposal distribution defined by a pretrained generator. The generator
provides a learned prior over discrete transformations, encoding coordinated
changes without explicitly enumerating edit-distance neighbourhoods. GenTuRBO
performs strongly across all three domains and achieves the highest mean final
scores on Ehrlich sequence optimisation and Sokoban coordination.

\end{abstract}

\raggedbottom
\section{Introduction}
\label{sec:introduction}

\begin{figure}[!t]
    \centering
    \includegraphics[width=0.90\textwidth]{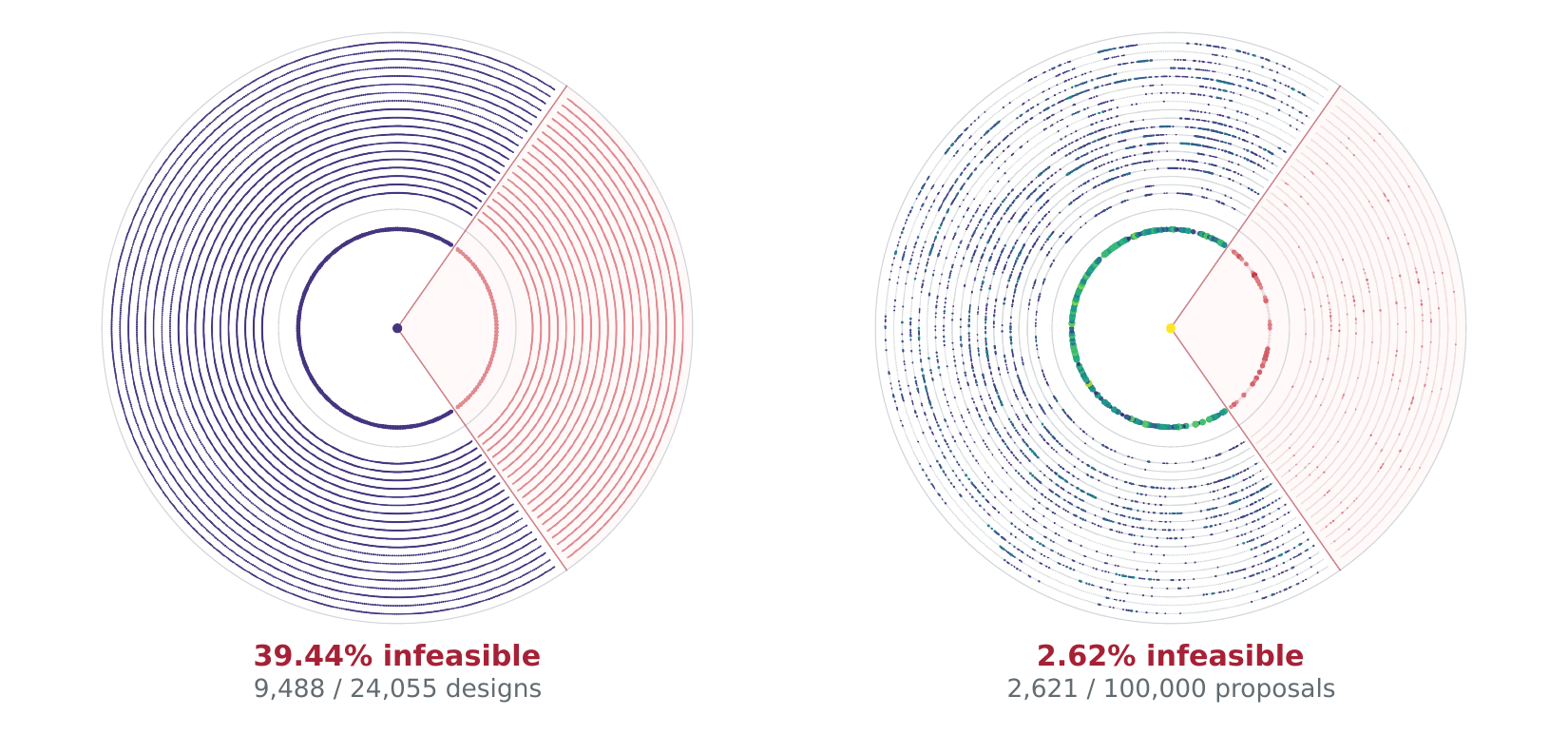}
    \caption{
\textbf{Generator-guided proposals concentrate probability on feasible sequences.}
Both panels show the same 24,055 sequences within two substitutions of a fixed
length-12 Ehrlich reference over a 20-token alphabet.
Each dot represents a distinct sequence: the centre is the reference, the inner
ring contains one-substitution variants, and the outer annulus contains
two-substitution variants.
\textbf{Left:} a uniform distribution over the Hamming neighbourhood assigns
39.44\% probability mass to infeasible sequences, rising to 59.63\% within four substitutions.
\textbf{Right:} 100,000 proposals from a generator trained on feasible sequences
assign only 2.62\% empirical probability mass to infeasible outcomes.
For each generator proposal, two positions are selected for corruption and
reconstructed.
Sequence locations are shared across panels, with infeasible sequences grouped
into a red sector for visual comparison. Marker colour and size on the right
indicate proposal frequency.
}
    \label{fig:visitations}
\end{figure}

Many scientific design problems require optimising structured discrete objects
using costly feedback. In protein engineering, for example, evaluating the
objective function for a sequence may require laboratory measurements or
expensive computation, while the number of possible designs is enormous. Such objects also
depend on coordinated choices: changing one residue can disrupt interactions
elsewhere in a protein, just as moving a wall or box can make a Sokoban board
unsolvable. The challenge is therefore to find designs with high objective values using as few
evaluations as possible while navigating these structural dependencies.

Bayesian optimisation (BO) addresses limited evaluation budgets by fitting a
probabilistic surrogate to observed objective values and using an acquisition function
to select promising evaluations. In high-dimensional spaces, however, both
surrogate modelling and acquisition optimisation become increasingly
difficult. Local BO methods mitigate these challenges by restricting search
to a smaller region around promising observed designs. TuRBO
\citep{NEURIPS2019_6c990b7a} implements this principle in continuous spaces
using an adaptive geometric trust region around the best observed design,
expanding the region after repeated improvements and contracting it after
repeated failures.

In discrete spaces, a natural analogue of a geometric trust region is an
edit-distance neighbourhood around the incumbent. However, edit distance
measures only how many components differ, not whether those changes respect
dependencies within the object. Useful improvements may require coordinated
edits that lie beyond a small metric neighbourhood, while even modest edit
radii can contain enormous numbers of candidates, many of them structurally
implausible or invalid. Explicit feasibility filters can reject candidates
that violate known constraints, but many structural dependencies are difficult
to specify as direct rules and cannot be fully captured by such filters.

Pretrained generative models offer a way to incorporate structural
dependencies directly into proposal generation. We introduce
\textbf{GenTuRBO}, a local Bayesian optimisation method that uses a fixed
pretrained generator as a prior over transformations of discrete objects. At
each iteration, GenTuRBO adaptively corrupts a promising design and uses the
generator to sample new designs consistent with the uncorrupted context.
The amount of corruption controls the scale of the resulting proposal distribution, providing a
stochastic discrete analogue of a continuous trust region. Rather than
treating all candidates within an edit-distance neighbourhood equally, the
generator concentrates proposal probability on coordinated changes that are
consistent with structure learned from data. The generator therefore
supplies a prior over plausible transformations, while the Bayesian
optimisation surrogate models the task-specific objective function and selects which
proposals to evaluate. Figure~\ref{fig:visitations} illustrates the
resulting proposal distribution on constrained Ehrlich sequences, where the
generator concentrates probability on a non-uniform subset enriched in
feasible designs.

Our contributions are:
\begin{enumerate}
\setlength{\topsep}{4pt}
\setlength{\itemsep}{3pt}
\setlength{\parsep}{0pt}
\setlength{\partopsep}{0pt}
\item \interlinepenalty=10000 We introduce a generator-defined trust region
for discrete Bayesian optimisation, using a fixed pretrained generative
model to provide a structured prior over local transformations.
\item We adapt the scale of this trust region from optimisation feedback by
controlling the amount of corruption applied before sampling new designs.
\item We evaluate GenTuRBO under matched evaluation budgets on constrained Ehrlich
sequences, protein fitness optimisation, and Sokoban coordination.
\end{enumerate}

\section{Preliminaries}
\label{sec:preliminaries}

\paragraph{Bayesian optimisation.}
Let $f:\mathcal{X}\rightarrow\mathbb{R}$ denote an expensive objective function and
$D_n=\{(x_i,y_i)\}_{i=1}^n$ the observations collected after $n$ evaluations,
where $y_i$ is a possibly noisy observation of $f(x_i)$.
Bayesian optimisation uses $D_n$ to construct a probabilistic surrogate of
the objective function and an acquisition function to select subsequent evaluations
\citep{mockus1989bayesian,garnett2023bayesian}.
We use a Gaussian process (GP) surrogate conditioned on $D_n$
\citep{frazier2018tutorial,rasmussen2006gaussian}.
For the best observed objective value $y_n^\star=\max_{i\leq n} y_i$, expected
improvement (EI) is
\[
    \alpha_{\mathrm{EI}}(x)
    = \mathbb{E}\left[\max\{f(x)-y_n^\star,0\} \mid D_n\right],
\]
where the expectation is taken under the GP posterior
\citep{jones1998efficient}.

\paragraph{Local Bayesian optimisation.}
\label{subsec:adaptivelocalbayesianoptimisation}
In continuous spaces, TuRBO \citep{NEURIPS2019_6c990b7a} restricts candidate
selection to an adaptive trust region centred on the incumbent $x^\star$,
the best observed design. For a $d$-dimensional domain, the trust region is
the hyperrectangle
\[
    \mathcal T = \left\{ x\in\mathcal X:
    \left|x^{(j)}-x^{\star(j)}\right| \leq \frac{L_j}{2},
    \quad j=1,\ldots,d \right\},
\]
where $x^{(j)}$ denotes the $j$-th coordinate of $x$ and $L_j$ is the
corresponding side length. TuRBO parameterises the region by a base length
$L$, rescaled using the GP lengthscales $\lambda_j$,
\[
    L_j = \frac{\lambda_j L}
    {\left(\prod_{k=1}^{d}\lambda_k\right)^{1/d}},
\]
preserving a trust-region volume of $L^d$. Repeated improvements expand $L$,
while repeated failures contract it, allowing the scale of local search to
adapt during optimisation.

\paragraph{Generative models.}
Let $t\in[0,1]$ parameterise a path between a reference distribution and clean
data \citep{lipman2023flow,campbell2024discreteflow}.
Given a clean object $x$, a corruption process samples an intermediate state
\[
    \tilde{x}\sim q_t(\cdot\mid x),
\]
where larger values of $t$ retain more information about $x$. For discrete
sequences, corruption can be defined through categorical transition kernels
that, for example, replace tokens uniformly or transition them to a masking
state \citep{austin2021structured}. A pretrained generative model with fixed
parameters $\psi$ then reconstructs a clean candidate $x'$ from
$(\tilde{x},t)$, defining
\[
    x' \sim P_\psi(\cdot\mid \tilde{x},t),
\]
corresponding to the learned reverse or generative process. Initialising
generation from an intermediate state produces variants conditioned on the
supplied object, with the corruption level controlling the amount of retained
information.

\paragraph{Ehrlich feasibility.}
\label{sec:Ehlrich}
Throughout the paper, we use the Ehrlich sequence-optimisation task
\citep{stanton2024closed,chen2025generalists} to motivate our work in a
controlled setting where structural constraints are explicit. The task is
designed to mimic aspects of biological sequence design, where feasibility
depends on complex dependencies between sequence elements rather than on
individual positions in isolation. In Ehrlich, a sequence is feasible if
every adjacent token pair is permitted by a fixed Markov transition matrix;
a single forbidden transition makes the sequence infeasible. This allows
structural feasibility to be assessed separately from the task-specific objective value.

\section{Challenges for Discrete TuRBO}
\label{sec:discrete-challenges}

Let $\mathcal X=\mathcal A^L$ denote the space of length-$L$ sequences over a
finite alphabet $\mathcal A$, and let
$d_H(x,x')=\sum_{\ell=1}^{L}\mathbf 1\{x_\ell\neq x'_\ell\}$ denote Hamming
distance. A direct discrete analogue of TuRBO restricts acquisition to the
neighbourhood
\[
    \mathcal T_n(k_n)
    = \{x\in\mathcal X:d_H(x,x_n^\star)\leq k_n\},
\]
where $x_n^\star$ is the incumbent and the integer radius $k_n$ increases or
decreases in response to optimisation success or failure. This preserves
adaptive local search, but creates two difficulties: small changes in radius
can produce large changes in neighbourhood size, and metric proximity does
not ensure structural feasibility.

\subsection{Edit-distance trust regions have coarse scales}
\label{sec:combinatorial}

In continuous TuRBO, the trust-region scale can vary continuously. A Hamming
neighbourhood, by contrast, changes only when its integer-valued radius
changes. For the Ehrlich setting ($L=256,\ |\mathcal A|=20$), a radius-one
neighbourhood contains 4,865 sequences, radius two contains approximately
11.8 million, and radius three approximately $1.9\times10^{10}$.
Increasing the radius from one to two therefore expands the candidate space
by more than three orders of magnitude, while reducing it below one leaves
only the incumbent. Edit radius consequently provides a coarse mechanism
for controlling the scale of local search. The corresponding counting
expressions are given in Supplementary Section~\ref{app:hamming-neighbourhood-size}.

\subsection{Edit-distance locality does not ensure feasibility}
\label{sec:feasibility}
In structured discrete domains, metric proximity alone need not imply that a design respects the dependencies governing feasibility or usefulness.
A single substitution may be invalid unless
accompanied by compensating changes elsewhere, so feasible improvements can
require several coordinated edits. Increasing the edit radius makes such
changes accessible, but also introduces a rapidly growing number of
candidates that violate the underlying constraints.

\begin{figure}[!t]
    \centering
    \includegraphics[width=0.55\textwidth]{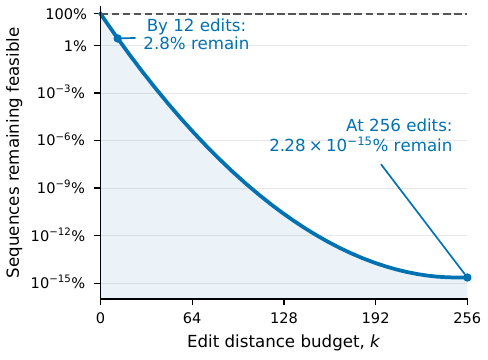}
    \caption{
\textbf{Feasibility collapses with edit radius.}
Percentage of length-$256$ sequences over a $20$-token alphabet within $k$
substitutions of one fixed feasible reference that satisfy the Ehrlich
transition rules, shown on a logarithmic vertical axis.
Counts are computed exactly (Supplementary Section~\ref{app:combinatorial-counts}). Only $2.8\%$ of
candidates remain feasible by twelve edits, and at $k=256$ --- where the
neighbourhood covers the entire sequence space --- the feasible fraction is
$2.28\times10^{-15}\%$.
}
    \label{fig:combined-constraints}
\end{figure}

One approach is to generate candidates from the edit-distance neighbourhood
and reject those that fail an explicit feasibility test. In the Ehrlich
task, however, only 2.8\% of sequences within twelve substitutions of a
fixed feasible reference satisfy the Markov transition constraints
(Figure~\ref{fig:combined-constraints}). Uniform proposal generation at that
radius would therefore spend most of its effort producing candidates that
are subsequently discarded. Moreover, many structured design problems do not
provide an explicit feasibility rule at all. These limitations motivate a
proposal mechanism that incorporates structural dependencies during
generation rather than only filtering candidates afterwards.

\section{GenTuRBO}
\label{sec:method}

GenTuRBO retains TuRBO's adaptive control of locality but replaces its
geometric trust region with a proposal distribution induced by a fixed
pretrained generator. Given an incumbent $x_n^\star$ and corruption level
$t_n$, candidates are generated in two stages:
\[
    \tilde{x}\sim q_{t_n}(\cdot\mid x_n^\star),
    \qquad x'\sim P_\psi(\cdot\mid\tilde{x},t_n).
\]
Marginalising over the corrupted state defines the proposal kernel
\[
    K_{\psi,t_n}(x'\mid x_n^\star)
    = \sum_{\tilde{x}} P_\psi(x'\mid\tilde{x},t_n)\,
    q_{t_n}(\tilde{x}\mid x_n^\star).
\]
For a fixed incumbent, this kernel defines a conditional distribution over
reconstructed candidates, which we call the \textbf{generative trust region}.
We parameterise its locality by the noising depth $r_n=1-t_n$: stronger
corruption retains less information about the incumbent and generally
produces broader proposals, while weaker corruption retains more information
and concentrates proposals locally.

Drawing $M$ candidates from $K_{\psi,t_n}(\cdot\mid x_n^\star)$ produces the
proposal pool $C_n$. A GP surrogate fitted to observed objective values in $D_n$
selects a batch of eligible candidates from $C_n$ with the largest marginal
expected improvement (EI) values for objective function evaluation.
The generator remains fixed throughout optimisation; new observations update
the surrogate, incumbent, and adaptive locality controller.
Algorithm~\ref{alg:generator-guided-trust-region} summarises the procedure;
Supplementary Section~\ref{app:turbo-genturbo} compares GenTuRBO with continuous TuRBO.

\begin{algorithm}[H]
\small
\caption{GenTuRBO}
\label{alg:generator-guided-trust-region}
\renewcommand{\algorithmicrequire}{\textbf{Input:}}
\renewcommand{\algorithmicensure}{\textbf{Output:}}
\begin{algorithmic}[1]
\Require Initial archive $D_0$, fixed pretrained generator $P_\psi$,
objective evaluation budget $N$, batch size $q$, proposal-pool size $M$, initial noising
depth $r_{\mathrm{init}}$, locality parameters
$r_{\min},r_{\max},\gamma_{\mathrm{expand}},\gamma_{\mathrm{shrink}},\epsilon_{\mathrm{imp}}$.
\State Initialise $n\gets |D_0|$, $D_n\gets D_0$, and $r\gets r_{\mathrm{init}}$.
\While{$n<N$}
    \State Set incumbent $x^\star\gets\arg\max_{(x,y)\in D_n}y$.
    \State Set corruption level $t\gets 1-r$.
    \State Fit GP surrogate to $D_n$.
    \State Sample $M$ candidates from $K_{\psi,t}(\cdot\mid x^\star)$ to form proposal pool $C_n$.
    \State Select batch $B_n\subseteq C_n$ of up to $\min\{q,N-n\}$ eligible candidates with the largest marginal EI values.
    \State Evaluate $f(x)$ for $x\in B_n$.
    \State Update archive
    \[
        D_{n+|B_n|}\gets D_n\cup\{(x,f(x)):x\in B_n\}.
    \]
    \State Update $r$ from the success or failure of the evaluated batch.
    \State Set $n\gets n+|B_n|$.
\EndWhile
\Ensure Best evaluated design $x^\star=\arg\max_{(x,y)\in D_n}y$.
\end{algorithmic}
\end{algorithm}

\subsection{Adaptive locality control}

An iteration is considered successful when the best objective value in the newly
evaluated batch improves on the previous incumbent by more than
$\epsilon_{\mathrm{imp}}$, where $\epsilon_{\mathrm{imp}}\geq0$ is the minimum
improvement required to count as a success. Because we use success and
failure thresholds of one, the noising depth is updated after every
evaluated batch. Following a success, $r$ is multiplied by
$\gamma_{\mathrm{expand}}>1$, up to a maximum $r_{\max}$; following a
failure, it is multiplied by $0<\gamma_{\mathrm{shrink}}<1$, down to a
minimum $r_{\min}$.

If an iteration fails while $r=r_{\min}$, we reset the noising depth to
$r_{\mathrm{init}}$. This serves a similar purpose to TuRBO's restart when
a trust region becomes too small, but GenTuRBO does not discard the
accumulated observations or initialise a new local model. Instead, the
archive and surrogate are retained and only the locality scale is reset.

\section{Related Work}
\label{sec:related-work}

\noindent\textbf{Discrete and combinatorial Bayesian optimisation.}
Existing approaches to discrete BO address high dimensionality through specialised kernels, trust regions, or low-dimensional representations. COMBO \citep{oh2019combo} models categorical and ordinal variables using a diffusion-kernel GP on a Cartesian-product graph, while CASMOPOLITAN \citep{pmlr-v139-wan21b} combines specialised kernels with Hamming-distance trust regions. Bounce \citep{papenmeier2023bounce} performs trust-region search in nested random embeddings of increasing dimension. GenTuRBO follows the principle of adaptive local search but replaces metric-based locality with the conditional proposal distribution of a fixed pretrained generator.

\noindent\textbf{Latent-space Bayesian optimisation.}
A common approach to structured optimisation is to embed discrete objects
in a continuous latent space and perform Bayesian optimisation there.
Weighted retraining adapts the learned representation towards high-scoring
observations \citep{NEURIPS2020_81e3225c}, while LaMBO jointly trains a
denoising autoencoder and multi-task GP for biological sequence design
\citep{stanton2022lambo}. COWBOYS decouples the generator and surrogate and
combines their information through a Bayesian update \citep{moss2025return}.
These methods optimise latent representations
rather than edits to the objects themselves; the resulting sequence or
structural changes are determined only after decoding. GenTuRBO instead
retains object-space incumbents and uses the pretrained generator only to
define local proposal distributions around them.

\noindent\textbf{Conditional noise--denoise generation.}
Several generative methods condition generation on an existing object by
partially corrupting it and then reconstructing from the resulting
intermediate state. SDEdit adds noise to a guide image before denoising,
with the noising level controlling how strongly the output follows the guide
\citep{meng2022sdedit}. RePaint similarly uses a fixed unconditional
diffusion model for image inpainting while preserving observed regions
during denoising \citep{lugmayr2022repaint}, and RFdiffusion uses partial
diffusion to generate structural variants of an existing protein backbone
\citep{vazqueztorres2024binders}. Across these methods, the corruption level
controls the degree of deviation from the supplied object. GenTuRBO
repurposes this mechanism as an adaptive locality controller for black-box
optimisation, updating the corruption level from optimisation progress
while keeping the pretrained generator fixed.

\noindent\textbf{Evolutionary and generator-guided search.}
Evolutionary methods generate candidates using hand-designed mutation and
crossover operators, including graph-based genetic algorithms for molecular
design \citep{jensen2019graph}, while BO-EVO combines evolutionary proposals
with surrogate-guided selection for protein engineering
\citep{hu2023protein}. Learned generators provide an alternative proposal
mechanism: guided discrete diffusion steers protein generation using an
acquisition function \citep{gruver2023protein}, and amortised Bayesian
optimisation trains a generator to propose acquisition-optimal candidates
directly \citep{swersky2020amortized}. GenTuRBO separates proposal structure
from task-specific optimisation: a fixed pretrained generator defines the
local candidate distribution, while the surrogate evaluates completed
proposals and selects which candidates receive objective function evaluations.
The separation is useful because structural plausibility and predicting task-specific
objective values are distinct modelling problems. A surrogate may accurately predict
the property being optimised without encoding whether a candidate is itself
plausible or viable.

\section{Experiments}
\label{sec:experiments}

We evaluate GenTuRBO on constrained Ehrlich sequences, protein fitness, and
Sokoban coordination under matched oracle budgets. In protein design, the
requirements for biological function are difficult to specify explicitly,
and high predicted fitness alone does not establish that proposed sequences
remain functional without experimental validation \citep{kirjner2024improving}.
We therefore complement the protein proxy benchmarks with Ehrlich and Sokoban,
where feasibility is explicitly defined and can be verified computationally.
In these controlled tasks, generators are trained on feasible examples,
allowing us to examine whether learned structural information helps preserve
feasibility while searching for higher-reward designs.

\subsection{Experimental protocol}
\label{sec:experimental-protocol}

Each reported condition uses 20 seeds and a budget of 2,000 charged oracle
evaluations, including initialisation. We report mean $\pm$ sample standard
deviation across seeds, with performance measured by the best score found
within the evaluation budget. The pretrained generator is not retrained
during GenTuRBO optimisation. Task-specific settings, baseline
implementations, component analyses, and significance tests are reported
in the Supplementary Material.

\subsection{Ehrlich sequence optimisation}
\label{sec:results-ehrlich}

\begin{figure}[!t]
    \centering
    \includegraphics[width=0.55\textwidth]{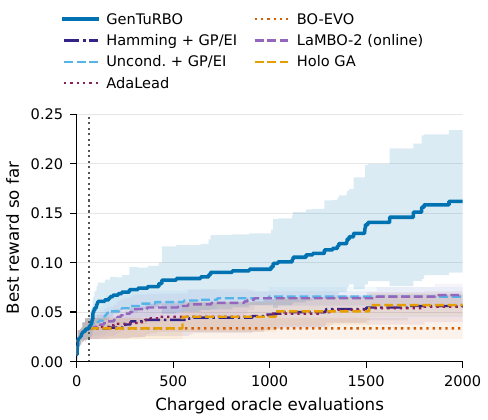}
    \caption{\textbf{Ehrlich optimisation.} Mean best-so-far reward with
    one sample standard deviation across 20 seeds. The dotted line marks
    the shared 64-evaluation initial archive. Holo GA is recorded at
    completed-generation checkpoints. Hamming + GP/EI uses direct mutation
    proposals screened by the spaced-pair GP/EI policy. AdaLead uses
    root-anchored rollouts and predicted-mean selection.}
    \label{fig:ehrlich-trajectories}
\end{figure}

The Ehrlich task \citep{stanton2024closed,chen2025generalists} rewards four
spaced motifs in length-256 sequences over a 20-token alphabet. Each motif
contains eight tokens, and the reward multiplies their quantised match
scores. The generator is trained without reward information on feasible
length-128 sequences and applied convolutionally at length 256. All iterative
methods share 64 initial evaluations within each seed, and GenTuRBO
subsequently selects batches of eight from 128 generated proposals. The GP
uses a sequence kernel based on counts of spaced token pairs, analogous to
$k$-spaced pair representations used in biological sequence modelling
\citep{chen2007flexible}. We compare GenTuRBO against Holo GA
\citep{stanton2024closed}, AdaLead \citep{sinai2020adalead}, BO-EVO
\citep{hu2023protein}, and online LaMBO-2 \citep{gruver2023protein}.

Both adaptive noising and GP-guided proposal selection improve Ehrlich
performance in the component analysis (Supplementary Section~\ref{app:component-ehrlich}).

GenTuRBO achieves the highest mean final reward under the matched evaluation
budget (Figure~\ref{fig:ehrlich-trajectories}; Supplementary Table~\ref{tab:ehrlich-headline}).
It outperforms online LaMBO-2 in 19 of 20 seeds with one tie, and outperforms
AdaLead, BO-EVO and Holo GA in all 20 seeds.
As an additional locality control,
GenTuRBO outperforms Hamming + GP/EI in 19 of 20 paired seeds, with one tie.
Additional baseline comparisons and locality diagnostics are reported in
Supplementary Section~\ref{app:ehrlich-diagnostics}.

\subsection{Protein fitness optimisation}
\label{sec:results-protein}

We optimise GFP and AAV on their Medium and Hard benchmark tasks
\citep{kirjner2024improving,sarkisyan2016local,bryant2021deep}. GenTuRBO uses
EvoDiff \citep{alamdari2023protein} as its fixed pretrained generator, with
fitness evaluated by a CNN proxy oracle and normalised using the corresponding
deep-mutational-scanning dataset. We do not assess the feasibility of the produced proteins, and therefore
scores may reflect high-scoring but unrealistic designs.
All methods share a fixed 128-sequence
initial archive for each task. We evaluate GenTuRBO with both Kermut
\citep{groth2024kermut} and one-hot Tanimoto
\citep{bajusz2015tanimoto,griffiths2023gauchelibrarygaussianprocesses} GP
configurations, and compare against Hamming + GP/EI, Holo GA
\citep{stanton2024closed}, AdaLead \citep{sinai2020adalead}, BO-EVO
\citep{hu2023protein}, and online LaMBO-2 \citep{gruver2023protein}.

\begin{table}[!t]
\centering
\footnotesize
\setlength{\tabcolsep}{1.5pt}
\renewcommand{\arraystretch}{1.05}
\caption{\textbf{Protein fitness optimisation.} Best normalised CNN score
after 2,000 charged evaluations, mean $\pm$ sample SD over 20 seeds.
Bold marks the highest reported mean; $^*$ denotes a positive mean
difference and $p<0.05$ against every other method in that task's table. There is no explicit feasibility check, so protein results may not correspond to valid designs.}
\label{tab:protein-headline}
\textbf{Medium tasks}\par\smallskip
\begin{tabular*}{0.62\textwidth}{@{\extracolsep{\fill}}lcc@{}}
\toprule
Method & GFP & AAV \\
\midrule
GenTuRBO (Kermut) & $0.983\pm0.074$ & $0.806\pm0.033$ \\
GenTuRBO (Tanimoto)$^\dagger$ & $0.854\pm0.149$ & $\mathbf{0.832\pm0.035}$ \\
Hamming + GP/EI & $\mathbf{1.161\pm0.067}^{*}$ & $0.824\pm0.041$ \\
Holo GA & $0.864\pm0.038$ & $0.656\pm0.028$ \\
AdaLead & $0.906\pm0.047$ & $0.802\pm0.053$ \\
BO-EVO & $0.624\pm0.007$ & $0.607\pm0.042$ \\
LaMBO-2 (online) & $0.775\pm0.035$ & $0.732\pm0.034$ \\
\bottomrule
\end{tabular*}
\par\medskip
\textbf{Hard tasks}\par\smallskip
\begin{tabular*}{0.62\textwidth}{@{\extracolsep{\fill}}lcc@{}}
\toprule
Method & GFP & AAV \\
\midrule
GenTuRBO (Kermut) & $0.845\pm0.069$ & $0.757\pm0.090$ \\
GenTuRBO (Tanimoto)$^\dagger$ & $0.933\pm0.081$ & $0.766\pm0.109$ \\
Hamming + GP/EI & $\mathbf{1.014\pm0.056}^{*}$ & $\mathbf{0.796\pm0.063}$ \\
Holo GA & $0.433\pm0.021$ & $0.653\pm0.028$ \\
AdaLead & $0.783\pm0.064$ & $0.722\pm0.025$ \\
BO-EVO & $0.317\pm0.061$ & $0.574\pm0.047$ \\
LaMBO-2 (online) & $0.685\pm0.050$ & $0.696\pm0.050$ \\
\bottomrule
\end{tabular*}
\par\smallskip\parbox{0.62\textwidth}{\footnotesize Hamming + GP/EI uses direct
Hamming mutation proposals with the one-hot Tanimoto GP and marginal EI.
$^\dagger$GFP-Medium GenTuRBO Tanimoto and Hamming + GP/EI fix noise variance to 0.001 in
standardised outcome units; other reported Tanimoto tasks learn noise.
Component comparisons: Supplementary Section~\ref{app:component-protein}.}
\end{table}

\begin{figure}[t]
    \centering
    \includegraphics[width=0.83\textwidth]{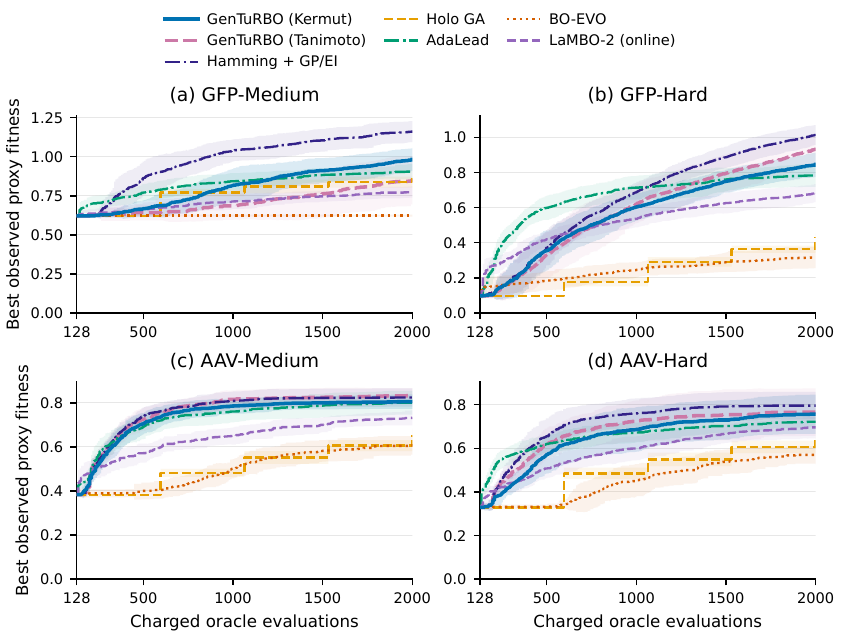}
    \caption{\textbf{Protein fitness optimisation.} Mean best-so-far
    normalised CNN score with one sample standard deviation across 20 seeds.
    The 2,000-label budget includes the shared 128-sequence archive.
    For GFP-Medium, GenTuRBO (Tanimoto) and Hamming + GP/EI fix likelihood
    noise variance to 0.001 in standardised outcome units; other Tanimoto
    tasks learn noise.
    Holo GA points mark completed generations. Hamming + GP/EI uses uniform
    exact-distance mutation proposals with an adaptive radius, a one-hot
    Tanimoto GP, and marginal EI on all four tasks.}
    \label{fig:protein-final}
\end{figure}

Hamming + GP/EI achieves the highest mean final score on GFP-Medium,
GFP-Hard, and AAV-Hard, while GenTuRBO with the Tanimoto surrogate achieves
the highest mean on AAV-Medium (Table~\ref{tab:protein-headline};
Figure~\ref{fig:protein-final}). Hamming + GP/EI's strong performance on the
learned protein fitness proxies contrasts sharply with its performance on the
explicitly constrained benchmarks. GenTuRBO outperforms it in 19 of 20 Ehrlich
runs, with one tie, and in all 20 Sokoban runs, where Hamming + GP/EI fails to
improve on the best board in the initial archive
(Figures~\ref{fig:ehrlich-trajectories} and~\ref{fig:sokoban-comparison}). However, unlike Ehrlich and Sokoban, these benchmarks
do not provide an independent measure of whether high-scoring proposals
satisfy the underlying biological constraints: the optimisation objective
itself is a learned predictor, and establishing whether proposed sequences
remain functional in practice would ultimately require experimental
validation. The exact-feasibility benchmarks therefore provide a complementary
test of whether a proposal mechanism preserves structural validity during
optimisation. Additional component, kernel, and diagnostic comparisons are
reported in Supplementary Sections~\ref{app:component-protein} and~\ref{app:protein-diagnostics}.

\subsection{Sokoban coordination optimisation}
\label{sec:results-sokoban}

\begin{figure}[t]
    \centering
    \includegraphics[width=\textwidth]{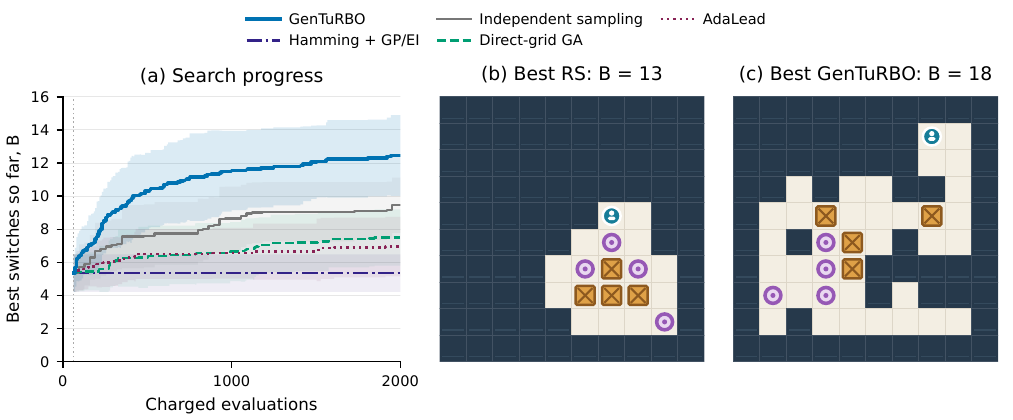}
    \caption{\textbf{Sokoban coordination optimisation.}
    (a) Best box-switch score found so far, averaged across 20 seeds.
    Shading shows one sample standard deviation. Each seed contributes its highest exactly
    scored $B$. Curves update at recorded batch checkpoints; the dotted line
    marks the shared 64-query initial archive.
    (b,c) Highest-scoring boards across the current 20-seed cohorts at
    2,000 queries: independent sampling ($B=13$) and GenTuRBO ($B=18$), both independently verified.
    Amber squares are boxes, purple circles are goals, and blue marks the
    player. Search implementation details are reported in Supplementary Section~\ref{app:sokoban}.}
    \label{fig:sokoban-comparison}
\end{figure}

We optimise $10\times10$ Sokoban boards containing four boxes and four goals.
The objective $B$ is the minimum number of switches between pushed boxes
over all solutions; the first pushed box does not count as a switch
\citep{jarusek2010difficulty}. GenTuRBO uses a reward-blind masked Transformer
following \citet{baghal2026solvable}, trained on 900,000 Boxoban boards
\citep{boxobanlevels}, as its fixed generator. Methods share 64 initial boards
within each seed and use the same bounded solver. Every oracle query consumes
budget, but only exactly scored solvable boards update the best-so-far score.
Search settings, baseline implementations, and solver details are reported
in Supplementary Section~\ref{app:sokoban}.

GenTuRBO achieves the highest mean final best score among the compared methods
(Figure~\ref{fig:sokoban-comparison}; Supplementary Table~\ref{tab:sokoban-headline}). It outperforms
direct-grid GA and Hamming + GP/EI in all 20 seeds, and AdaLead in 19 of 20
seeds with one tie.

The Hamming + GP/EI control shares GenTuRBO's parent, surrogate, acquisition
function, and locality controller, but replaces generator reconstruction with
direct Hamming mutations. It finds no new solvable boards, whereas GenTuRBO
retains 88.76\% solvability across charged queries. The controlled comparison
supports the value of generator-guided proposals for maintaining solvability
during optimisation. The highest-scoring GenTuRBO board requires 18 box
switches, independently verified by uniform-cost search. Full diagnostic
results are reported in Supplementary Section~\ref{app:component-sokoban}.

\section{Conclusion}
\label{sec:conclusion}

We introduced GenTuRBO, a discrete Bayesian optimisation framework that
replaces geometric trust regions with adaptive proposal distributions defined
by a fixed pretrained generator. By combining generator-guided
transformations, adaptive corruption levels, and surrogate-based candidate
selection, GenTuRBO enables structured local search without relying on
explicit edit-distance neighbourhoods.

GenTuRBO achieved the highest mean final performance among budget-matched
methods on Ehrlich sequences and Sokoban coordination. Direct Hamming + GP/EI
achieved the highest mean proxy fitness on three of the four protein tasks,
while GenTuRBO (Tanimoto) led on AAV-Medium. Across these benchmarks,
generator-defined locality shows its clearest advantage when structural
constraints can be evaluated directly, whereas direct Hamming search is most
competitive on the learned protein proxy objectives.

While our evaluation focuses on discrete spaces, the effectiveness of
GenTuRBO also depends on the coverage of the pretrained generator.
Investigating this dependence and extending generator-defined trust regions
to continuous and mixed search spaces are promising directions for future
work.

\subsection*{Code and data availability}

Source code, experiment configurations, dependency locks and reproduction
instructions are available at
\url{https://github.com/Detriatis/GenTuRBO}.
The repository also links to the companion reproduction bundle
of frozen data, checkpoints and numerical reference results, together with
its checksum and import instructions.

\subsection*{AI use statement}

The project idea, experimental design, and interpretation of results were
developed without the assistance of generative AI. Generative AI tools
assisted with most software implementations used in this study, as well as
manuscript drafting and editing and figure-generation code. AI-generated
code was manually reviewed and tested for consistency with the intended
implementations. The authors take responsibility for the final code,
reported results, and manuscript.

\subsection*{Acknowledgements}

This work was supported by the Engineering and Physical Sciences Research
Council [EP/Y030826/1].

The authors acknowledge the use of resources provided by the Isambard-AI
National AI Research Resource (AIRR). Isambard-AI is operated by the
University of Bristol and is funded by the UK Government's Department for
Science, Innovation and Technology (DSIT) via UK Research and Innovation;
and the Science and Technology Facilities Council [ST/AIRR/I-A-I/1023].
For a description of the facility, see \citet{mcintoshsmith2024isambard}.

\bibliography{references}

\clearpage
\begin{center}
\phantomsection
\pdfbookmark[1]{Supplementary Material}{supplementary-material}
{\LARGE Supplementary Material\par}
\end{center}
\vspace{1ex}

\setcounter{section}{0}
\setcounter{figure}{0}
\setcounter{table}{0}
\renewcommand{\thesection}{S\arabic{section}}
\renewcommand{\thefigure}{S\arabic{figure}}
\renewcommand{\thetable}{S\arabic{table}}
\renewcommand{\theHsection}{S\arabic{section}}
\renewcommand{\theHfigure}{S\arabic{figure}}
\renewcommand{\theHtable}{S\arabic{table}}

\section{Experimental Protocol}
\label{app:experimental-setup}

The main comparisons use 20 search seeds and 2,000 charged oracle evaluations
per run, including initial labels. Means and error bands use sample standard
deviations across seeds. The initial archive is shared within each Ehrlich or
Sokoban seed; each protein task instead uses one fixed archive across all
methods and seeds. Curves and endpoints report each run's best observed
score so far, averaged across seeds; Sokoban uses only exact switch scores.
Main comparison curves hold the incumbent constant between recorded batch
or generation checkpoints. Proposal generation, duplicate handling, surrogate
fitting and acquisition do not consume oracle calls. Ehrlich feasibility and
Sokoban validity/solvability are assessed only as part of a charged query;
selected invalid designs still consume budget. The privileged oracle selector
is a separate diagnostic, described below. GenTuRBO keeps its generator fixed
throughout search. Table~\ref{tab:experimental-protocol} summarises the
executed configurations; Ehrlich baseline operators are in Section~\ref{app:ehrlich-baselines},
and the Sokoban objective, solver and operators are in Section~\ref{app:sokoban}.

GenTuRBO's GP uses the complete available paid history rather than only local
observations. Batch selection takes the eligible proposals with the largest
marginal EI values, without local penalisation. Censored Sokoban outcomes
are excluded from GP labels, as detailed in Section~\ref{app:sokoban}.

\begin{table}[H]
\centering
\footnotesize
\setlength{\tabcolsep}{4pt}
\renewcommand{\arraystretch}{1.15}
\caption{\textbf{Configurations for the main GenTuRBO comparisons.}
Pool and batch sizes describe generator variants; baseline exceptions are
documented separately. GFP-Medium Tanimoto uses fixed likelihood noise
variance 0.001 in standardised outcome units; other displayed protein
configurations use learned noise.}
\label{tab:experimental-protocol}
\begin{tabular}{@{}p{0.18\textwidth}p{0.23\textwidth}p{0.27\textwidth}p{0.23\textwidth}@{}}
\toprule
Setting & Ehrlich & Protein & Sokoban \\
\midrule
Design space & 256 tokens; alphabet 20 & GFP: 237 residues; AAV: 28-residue window & $10\times10$ grid; four boxes and goals \\
Frozen generator & Residual CTMC CNN & EvoDiff & Masked Transformer \\
Initial / online labels & 64 / 1,936 & 128 / 1,872 & 64 / 1,936 \\
Proposal pool & 128 & GP: 256; Random: 24 & 128 \\
Acquired batch & 8 & 8 & 8 \\
Conditional parent & Best observed sequence & Best observed sequence & Best exactly scored solvable board \\
GP kernel & Spaced pairs, window 32 & Kermut; one-hot Tanimoto comparison & One-hot Tanimoto \\
Acquisition & Top-eight marginal EI & Top-eight marginal EI & Top-eight marginal EI \\
GP history & Complete paid archive & Complete paid archive & Complete finite paid history \\
GP refitting & Every eight paid batches & Every eight batches & Every eight archive updates \\
Adaptive coordinate & Requested edits 2--128; initial 64 & Noising depth; Table~\ref{tab:protein-tr-hyperparameters} & Masked cells 5--40; initial 20 \\
Hamming GP/EI pool / batch & 128 / 8; spaced-pair GP & 256 / 8; Tanimoto GP & 128 / 8; Tanimoto GP \\
Hamming edit controller & 2--128; initial 64 & 1--$L$; initial $L/4$ (GFP), $L/2$ (AAV) & 5--40; initial 20 \\
Endpoint & Best motif reward & Best normalised CNN score & Best minimum box-switch count \\
\bottomrule
\end{tabular}
\end{table}

\subsection{Ehrlich task and generator}
\label{app:ehrlich-protocol}

We use the Ehrlich benchmark of \citet{stanton2024closed}, also studied by
\citet{chen2025generalists}, with length $L=256$, alphabet size 20, $C=4$
motifs, motif length $K=8$ and quantisation $Q=4$. A sequence is feasible
when all adjacent transitions have positive probability under the task's
Markov matrix $T$. For motif tokens $m_j^{(i)}$ at strictly increasing offsets
$s_j^{(i)}$ with $s_1^{(i)}=0$, define
\[
M_i(x)=\max_{1\leq\ell\leq L-s_K^{(i)}}
\sum_{j=1}^{K}\mathbf{1}\{x_{\ell+s_j^{(i)}}=m_j^{(i)}\},
\qquad
h_Q(M)=\frac{1}{Q}\left\lfloor\frac{QM}{K}\right\rfloor.
\]
The executed reward is $\prod_{i=1}^{C}h_Q(M_i(x))$ for feasible sequences
and zero otherwise. Every selected infeasible sequence consumes an oracle
call. The product requires progress on every motif: a zero motif score makes
the complete reward zero.

The generator is trained without rewards, motif identities or motif offsets
on valid length-128 sequences from the Markov law. Its 256-dimensional token
embeddings receive a timestep embedding and pass through five residual
Conv1D blocks (kernel 9, dilation 1, SiLU, dropout 0.1), followed by a
$1\times1$ vocabulary projection. There are no length-specific parameters;
the same checkpoint is applied convolutionally at length 256. Training uses
a uniform CTMC interpolant: a token is retained with probability $t$ and
otherwise replaced uniformly. Cross-entropy predicts clean tokens, with
$t\sim\mathrm{Beta}(2,1)$, AdamW learning rate $2\times10^{-4}$ and weight
decay 0.01. The search checkpoint is the 18,000-update checkpoint, with
training batch size 128 and training seed 0, shared by all search seeds.

For token-wise corruption, the noising depth can equivalently be expressed
as a nominal corruption budget $e_n$, with
\[
    t_n = 1-\frac{e_n}{L}.
\]
Here, $e_n=L(1-t_n)$ is the expected number of positions selected for
corruption. The realised parent--child edit distance is determined after
reconstruction and need not equal $e_n$.

Conditional generation uses one best archive sequence and 128 proposals at
a common noising time. Uniform corruption may redraw the current token,
and subsequent denoising can alter the realised distance.
The CTMC uses a 20-interval logarithmic time grid concentrated toward the
clean endpoint; conditional rollouts traverse the remaining grid suffix. The random-time arms draw one
$t\sim\mathrm{Unif}(0,1)$ per proposal pool. After removing duplicates and
previously scored sequences, GP selection takes the eight largest marginal
EI values, without sequential fantasy updates or local penalisation.

The spaced-pair GP fixes its outputscale to one and fits the constant mean
and likelihood noise on the schedule in Table~\ref{tab:experimental-protocol}.
The edit budget expands by 1.5 after an improvement exceeding $10^{-6}$,
shrinks by 0.75 otherwise, and resets to 64 after failure at the lower bound.
Requested edits are converted to noising depth; realised parent--child
distance is measured separately in Section~\ref{app:ehrlich-locality}.

\subsection{Protein benchmarks and oracle}
\label{app:protein-protocol}

The GFP and AAV tasks follow the limited-data benchmark construction of
\citet{kirjner2024improving}, using the DMS datasets of
\citet{sarkisyan2016local} and \citet{bryant2021deep}. Medium initial-data
subsets lie between the 20th and 40th fitness percentiles; Hard subsets lie
below the 30th percentile. Eligible sequences must also be at least six
(Medium) or seven (Hard) edits from every sequence in the top-fitness region,
defined by the 99th percentile. Distances use the original benchmark's edit
distance calculation.

\begin{table}[H]
\centering
\small
\caption{\textbf{Protein initial-data subsets.} Published counts are
benchmark reference values; executed rows and unique sequences describe the
input CSV subsets used in these experiments. Each search starts from the
fixed top-128 archive within its subset. Fitness ranges are those of the
benchmark subsets.}
\label{tab:protein-initial-subsets}
\begin{tabular}{lrrrrrr}
\toprule
Task & Length & \shortstack{Published\\count} & \shortstack{Executed\\rows} & \shortstack{Unique\\sequences} & \shortstack{Fitness\\range} & \shortstack{Minimum\\edits} \\
\midrule
GFP-Medium & 237 & 2,828 & 2,828 & 2,828 & $[0.01,0.62]$ & 6 \\
GFP-Hard & 237 & 2,426 & 2,426 & 2,426 & $[0.00,0.10]$ & 7 \\
AAV-Medium & 28 & 2,139 & 2,145 & 2,111 & $[0.29,0.38]$ & 6 \\
AAV-Hard & 28 & 3,448 & 3,453 & 3,347 & $[0.00,0.33]$ & 7 \\
\bottomrule
\end{tabular}
\end{table}

The archive manifest at
\texttt{benchmarks/gfp\_aav/initial\_archives/manifest.json} identifies each
\texttt{*\_top128} JSONL archive by its sequences, scores and dataset indices.
Archives are not resampled between methods or seeds. The terminal oracle is a CNN fitness
predictor, so these are online in-silico optimisation experiments. If
$\mathcal O_\tau(x)$ is its raw prediction, the reported score is
\[
f_\tau(x)=\frac{\mathcal O_\tau(x)-y_{\min,\tau}}
{y_{\max,\tau}-y_{\min,\tau}},
\]
where the extrema come from the full task DMS dataset. Predictions are not
clipped to $[0,1]$. Optimisers access the initial labels and subsequent oracle
answers, not the remaining DMS labels.

We use the frozen 38M-parameter uniform-transition EvoDiff-D3PM model
\citep{alamdari2023protein}. Its native forward and reverse transition
matrices implement corruption and denoising; the reverse chain traverses
all native steps from the chosen noise level, without timestep subsampling.
Each conditional pool starts from the best observed sequence. Proposals use the standard
20-amino-acid alphabet and fixed task length; duplicate proposals and
previously evaluated sequences are removed before acquisition. RT draws
$t\sim\mathrm{Unif}(0,1)$; TR adapts noising depth $r=1-t$ with the settings
in Table~\ref{tab:protein-tr-hyperparameters}. The Random and GP arms use
different proposal-pool sizes (Table~\ref{tab:experimental-protocol}).

\begin{table}[H]
\centering
\small
\caption{\textbf{Protein adaptive-noising parameters.} The controller updates
$r=1-t$; the stored parameter \texttt{tr\_mean} is $t_{\rm init}$.
The two difficulty levels use the same settings within each protein.}
\label{tab:protein-tr-hyperparameters}
\begin{tabular}{lrrrrrrr}
\toprule
Task & $t_{\min}$ & $t_{\max}$ & $t_{\rm init}$ & $\gamma_{\rm expand}$ & $\gamma_{\rm shrink}$ & $\tau_s$ & $\tau_f$ \\
\midrule
GFP-Medium / Hard & 0.0 & 0.99 & 0.75 & 1.5 & 0.75 & 1 & 1 \\
AAV-Medium / Hard & 0.0 & 0.95 & 0.50 & 1.5 & 0.75 & 1 & 1 \\
\bottomrule
\end{tabular}
\end{table}

\paragraph{Surrogates and acquisition.}
Protein GPs standardise outcomes and use a constant mean and fixed unit outer
outputscale. Table~\ref{tab:experimental-protocol} gives their archive,
refitting and marginal-EI batch policy. The Tanimoto implementation
uses flattened amino-acid one-hot features, a BoTorch \texttt{SingleTaskGP},
Gauche's Tanimoto kernel, a Gaussian likelihood and numerical jitter
$10^{-4}$ \citep{griffiths2023gauchelibrarygaussianprocesses}. It fits mean
and noise, except for GFP-Medium in the main comparison, whose noise variance
is fixed to 0.001 in standardised outcome units for the generator GP arms
and Hamming + GP/EI. AdaLead and BO-EVO retain learned Tanimoto noise on all
four protein tasks. The GFP-Medium fixed-noise choice followed five initial
GenTuRBO runs, retained alongside 15 subsequent seeds; the full 20-seed
comparison is therefore not wholly prospective with respect to that choice.
Hamming uses its own completed 20-seed fixed-noise cohort. Earlier
learned-noise runs are preserved separately and not pooled into these cohorts.
Marginal EI selects the eight highest individual values without fantasy
updates or local penalisation.

Kermut \citep{groth2024kermut} combines a structure kernel with an RBF kernel
over mean-pooled ESM-2 embeddings. ProteinMPNN probabilities and C-alpha
coordinates align to the mutation window; ESM embeddings use the full GFP or
AAV construct. Its mixture, structural decay and scale, and ESM lengthscale
are fitted alongside the constant mean and likelihood noise; its optional
zero-shot mean is not used. These are two complete surrogate configurations.

\paragraph{Direct Hamming control.}
Hamming + GP/EI changes the proposal operator while retaining the best-parent
policy, full-history Tanimoto GP, refitting schedule and marginal-EI batch
selection. It samples $k$ distinct sites uniformly and substitutes each with
a uniformly chosen non-current amino acid, giving exact Hamming distance $k$.
The adaptive edit coordinate is rounded to an integer for mutation, spans
$1$ to $L$, and starts at $L/4$ for GFP and $L/2$ for AAV. It expands by 1.5
after one improving batch and contracts by 0.75 after one non-improving batch,
with reset to its initial value after failure at the lower bound. These
bounds are in edits, whereas the EvoDiff controller acts on noising depth.

\paragraph{AdaLead.}
We use the FLEXS-derived root-anchored AdaLead implementation
\citep{sinai2020adalead} with the same algorithm parameters across proteins,
Ehrlich and Sokoban. Parents satisfy $f(x)\geq f_{\max}-0.05|f_{\max}|$;
the retained archive list is repeated or truncated in measured order to
eight roots. Each site is independently mutated with probability $1/L$
($\mu=1$), drawing from the complete alphabet, including the current symbol.
Recombination is disabled. Descendants continue a rollout while their
predicted mean is at least the fixed root's predicted mean. Each paid round
uses exactly 128 surrogate predictions, including repeated root predictions,
with configured evaluation chunk size 20 and the final root chunk truncated
to respect that cap. Eight novel candidates with the largest predicted means
are queried directly, without a second EI pass. The protein surrogate is the
full-history one-hot Tanimoto GP, with mean/noise refitted every eight rounds.
Mutation uses seeded PyTorch draws. Thus 128 AdaLead model predictions and
256 protein generator proposals are distinct screening workloads; the
charged oracle budget and paid batch size are matched.

\paragraph{Other external baselines.}
Holo GA uses the released discrete-evolution operators
\citep{stanton2024closed}, survival quantile 0.01, mutation and recombination
probabilities $1/L$, and four generations of 468 evaluations after the
128-label archive. Its trajectory updates at generation boundaries.
The adapted BO-EVO \citep{hu2023protein} draws parents from the latest measured
batch with probabilities proportional to $\exp(10f(x))$. It independently
mutates sites with probability $1/L$, conditioned on at least one mutation,
using non-current replacements and no crossover. For each of eight paid
slots, eight novel children are screened by $\mu(x)+0.2\sigma(x)$, for 64
surrogate predictions per round. Selected children become virtual parents;
a predicted standard deviation exceeding twice the first child's value
triggers resampling from the measured parent batch. Its surrogate is the
same full-history, learned-noise Tanimoto GP used by AdaLead.

Online LaMBO-2 \citep{gruver2023protein} starts from random model weights and
the shared 128-label archive, without a separate protein pretraining corpus.
It trains for one initial epoch and one epoch on the complete paid history
after each round, with training batch size 128. Each round directly returns
eight guided designs, with one design step, two mutations per step and a cap
of eight guidance updates. It uses its learned objective model and guided
acquisition rather than the shared GP/EI selector. BO-EVO and LaMBO-2 are
budget-matched adaptations; their warm starts and model training differ from
the original biological protocols. All methods receive the same paid oracle
budget, but proposal operators and screening workloads differ.

\paragraph{Supplementary controls.}
Ehrlich and protein unconditional generation starts cold, without the shared
initial archive. The myopic oracle selector scores its entire current pool
before choosing a batch. These extra scores make it a privileged diagnostic,
outside the matched-budget comparison; its separate trajectory is not an
upper bound on another policy. Section~\ref{app:protein-surrogate-checks}
reports the current-cohort protein noise and posterior-variance-floor checks.

\FloatBarrier
\section{Combinatorics of Constrained Hamming Neighbourhoods}
\label{app:combinatorial-counts}
\subsection{Metric neighbourhood size}
\label{app:hamming-neighbourhood-size}

For a reference $x\in\mathcal A^L$, let $\mathcal H_k(x)$ contain strings
at exactly $k$ substitutions and $\mathcal B_k(x)$ those within $k$.
Choosing the edited positions and their replacement symbols gives the counts
used in Section~\ref{sec:combinatorial}:
\[
 |\mathcal H_k(x)|=\binom{L}{k}(|\mathcal A|-1)^k,
 \qquad
 |\mathcal B_k(x)|=\sum_{j=0}^k\binom{L}{j}(|\mathcal A|-1)^j.
\]
For fixed $k$ and growing $L$, the outer shell dominates, giving leading
order $L^k(|\mathcal A|-1)^k/k!$.

\subsection{Sparse pairwise constraints}

For Ehrlich's fixed Markov transition matrix $T$, define the permitted-pair relation
$\mathcal R=\{(a,b):T_{ab}>0\}$ and its binary adjacency matrix $G$.
Write $\mathcal F_{\mathcal R}$ for the sequences whose adjacent token pairs
all belong to $\mathcal R$. A
path-counting argument gives
\[
    |\mathcal F_{\mathcal R}|
    =
    \boldsymbol 1^\top G^{L-1}\boldsymbol 1.
\]
If every symbol has exactly $d_{\mathcal R}$ permitted successors and every
symbol is allowed initially, there are $|\mathcal A|$ choices at the first
position and $d_{\mathcal R}$ choices at each of the remaining $L-1$
positions.  Consequently,
\[
    |\mathcal F_{\mathcal R}|
    =
    |\mathcal A|d_{\mathcal R}^{L-1},
    \qquad
    \frac{|\mathcal F_{\mathcal R}|}{|\mathcal A|^L}
    =
    \left(\frac{d_{\mathcal R}}{|\mathcal A|}\right)^{L-1}.
\]
When $d_{\mathcal R}<|\mathcal A|$, even this simple pairwise rule retains an
exponentially decreasing fraction of the ambient strings as $L$ grows.

\subsection{Exact constrained neighbourhood counts}

The cumulative neighbourhood-count figure in the main text
(Figure~\ref{fig:combined-constraints}) uses the saved
length-$256$, $20$-token
Ehrlich transition support $G_{ab}=\mathbf 1[T_{ab}>0]$ and the fixed feasible
reference with index $58$ in search seed $0$'s initial archive, fixed before
the counting analysis. All initial tokens are permitted. Motif
matching changes reward, not membership of the feasible set counted here.

Let $D_i(b,j)$ count feasible prefixes of length $i$ ending in token $b$ with
exactly $j$ substitutions relative to the first $i$ reference tokens. Set
\[
 D_1(b,j)=\mathbf 1[j=\mathbf 1\{b\ne x_1\}],
\]
and, treating out-of-range edit indices as zero, propagate
\[
 D_i(b,j)=\sum_{a\in\mathcal A}G_{ab}
 D_{i-1}\!\left(a,j-\mathbf 1\{b\ne x_i\}\right).
\]
Then $N_{\mathrm{feas}}(k)=\sum_{j=0}^k\sum_b D_L(b,j)$.
Each feasible sequence follows exactly one prefix path, so this recurrence
counts distinct strings once and handles both adjacent and boundary edits.
Integer arithmetic gives exact counts; logarithms are taken only for plotting.
The implementation agrees with exhaustive enumeration on small directed
transition systems, all one-edit neighbours of the saved reference, and the
unrestricted closed form when every transition is allowed. Summing all
shells also recovers $\boldsymbol 1^\top G^{L-1}\boldsymbol 1$.
The portable plotting script and saved support/reference data reproduce all
counts without loading a trained generator or evaluating the reward.

\subsection{Separated-edit approximation}

Now fix $x\in\mathcal F_{\mathcal R}$ and a set
$I\subseteq\{2,\ldots,L-1\}$ of $k$ pairwise non-adjacent edit positions.  For
each $i\in I$, let
\[
    c_i
    =
    \left|
        \left\{
            a\in\mathcal A\setminus\{x_i\}:
            (x_{i-1},a)\in\mathcal R,\,
            (a,x_{i+1})\in\mathcal R
        \right\}
    \right|.
\]
Because the affected pairwise constraints do not overlap, the exact number of
rule-compatible replacements at these fixed positions is
$\prod_{i\in I}c_i$.  This quantity depends on the incumbent and on the detailed
pair relation; regular out-degree alone does not determine it.

For a counting approximation, suppose instead that each new adjacent pair is
permitted independently with probability $\eta$, with randomness taken over
this stylised compatibility model.  Each interior edit changes two pairs, so a
candidate with $k$ separated edits survives with probability approximately
$\eta^{2k}$.  When $k^2/L\ll1$, boundary and adjacent edit sets form a small
share of the Hamming shell, giving
\[
    \mathbb E\!\left[
        |\mathcal H_k(x)\cap\mathcal F_{\mathcal R}|
    \right]
    \approx
    \binom{L}{k}(|\mathcal A|-1)^k\eta^{2k}.
\]
Equivalently, the expected rejected portion of the shell is approximately
$1-\eta^{2k}$.  The approximation is not exact for boundary edits, adjacent
edits, heterogeneous pairwise rules, or dependent constraints.

These calculations are illustrative rather than a model of biological or
chemical feasibility.  Real structural constraints can be long-range,
higher-order, task-dependent, and unknown.  They motivate a non-uniform
proposal prior, not the claim that the generator's support equals
$\mathcal F_{\mathcal R}$ or that generated candidates are guaranteed to be
feasible.

\subsection{Visualisation of generator visitation}
\label{app:visitation-visualisation}

The introductory visitation figure (Figure~\ref{fig:visitations}) uses one fixed length-12 Ehrlich reference
over a 20-token alphabet. Its radius-two Hamming neighbourhood contains
$N=24{,}055$ distinct sequences: the reference, 228 one-substitution variants,
and 23,826 two-substitution variants. Each sequence occupies the same location
in both panels. The centre is the reference, the inner ring contains
one-substitution variants, and the entire outer band contains two-substitution
variants. A sequence is infeasible if any adjacent token pair is forbidden by
the task's Markov chain. All infeasible sequences are grouped into a
110-degree red sector; its angle is a layout choice, not a representation of
probability mass.

The left panel assigns probability $1/N$ to every sequence. The right panel
uses 100,000 draws from the fixed pretrained generator with sampling seed
20260901. Each draw corrupts two uniformly selected distinct positions and
reconstructs only those positions. Reconstruction can restore either original
token, so proposals may differ from the reference at zero, one, or two
positions. This is a single-reference proposal illustration, not a full
optimisation comparison.

Faint points on the right retain the complete neighbourhood. Coloured
highlights require empirical probability at least $0.5/N$, equivalent to
at least three visits. The 3,843 highlighted sequences account for 95.047\%
of all draws; probabilities are not renormalised after this display cutoff.
Purple-to-yellow colours indicate increasing visitation frequency for feasible
sequences, and light-to-dark red does the same for infeasible sequences.
Larger neighbour markers emphasise more frequent proposals, not higher
rewards. The infeasible percentages use all outcomes, including those below
the cutoff and returns to the reference: $9{,}488/24{,}055=39.44\%$ under
uniform sampling and $2{,}621/100{,}000=2.62\%$ under generator sampling
(rounded to two decimal places).

\section{Ehrlich Mechanism and Surrogate Diagnostics}
\label{app:ehrlich-diagnostics}
\subsection{Absolute edit-budget calibration}
\label{app:ehrlich-locality}

We use two synthetic Ehrlich sequence-design conditions as controlled mechanism
diagnostics: \texttt{l128\_q4\_k4} and \texttt{l256\_q4\_k8}.  The latter
generates length-256 sequences with the convolutional checkpoint and evaluates
them under the length-256, motif-length-eight reward definition.  For each
condition, 20 source seeds begin with 64 unconditional evaluations and then
run 24 conditional batches.  Each conditional batch contains 128 raw
proposals, of which eight are selected uniformly at random for oracle
evaluation.  The controller requests an absolute Hamming edit budget between
2 and 128.  The offline diagnostic retains all raw proposals and separately
identifies the valid acquisition-eligible and selected subsets.

\begin{figure}[H]
    \centering
    \includegraphics[width=\textwidth]{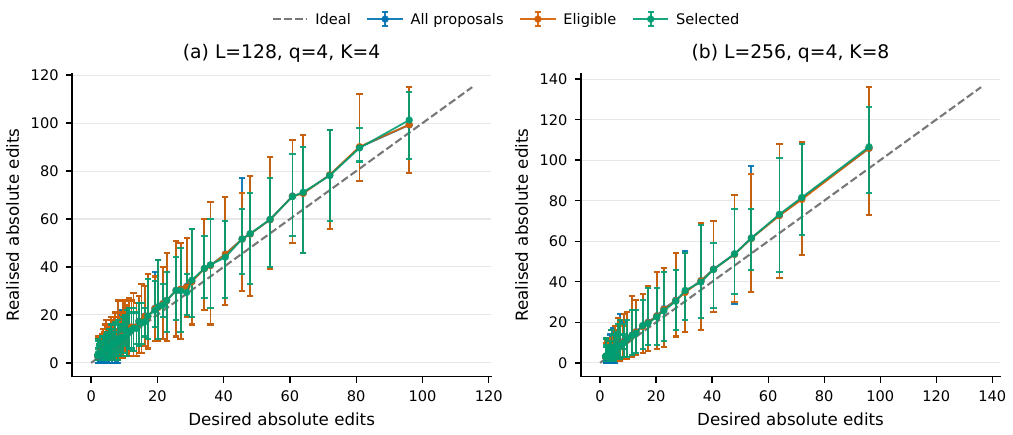}
    \caption{
    \textbf{Requested versus realised absolute edit distance on the Ehrlich
    mechanism diagnostic.}
    Points show mean parent--proposal Hamming distance at each requested edit
    budget for all proposals, valid acquisition-eligible proposals, and the
    eight randomly selected proposals per batch.  Error bars span the observed
    minimum and maximum within each requested-budget group; they are not
    confidence intervals.  The dashed diagonal is exact calibration.  Each
    panel aggregates 20 source seeds and 61,440 raw proposals.
    }
    \label{fig:ehrlich-generator-locality}
\end{figure}

Figure~\ref{fig:ehrlich-generator-locality} compares the requested budget with
the realised parent--proposal Hamming distance.  Across all 61,440 proposals
per condition, the association is strong (Pearson $r=0.976$ at $L=128$ and
$r=0.972$ at $L=256$).  The association is nearly unchanged for the randomly
selected proposals ($r=0.976$ and $r=0.973$, respectively), so it is not an
artefact of acquisition.  Table~\ref{tab:ehrlich-generator-locality}
summarises absolute calibration error.  Realised distances are systematically
slightly larger than requested on average, and the diagnostic does not show
that neighbourhoods are nested or that the same calibration transfers to a
different generator or scientific domain.  It establishes only that the
requested absolute edit budget is an effective locality-control coordinate in
these two Ehrlich conditions.

\begin{table}[H]
\centering
\small
\caption{
\textbf{Absolute edit-budget calibration.}
Values aggregate 20 source seeds.  MAE is the mean absolute difference between
the requested and realised parent--proposal Hamming distance.}
\label{tab:ehrlich-generator-locality}
\begin{tabular}{llrrr}
\toprule
Condition & Population & $n$ & Realised edits & MAE \\
\midrule
$L=128,Q=4,K=4$ & All & 61,440 & 28.66 & 4.69 \\
 & Eligible & 55,498 & 28.62 & 4.67 \\
 & Selected & 3,840 & 28.62 & 4.66 \\
\midrule
$L=256,Q=4,K=8$ & All & 61,440 & 27.65 & 5.00 \\
 & Eligible & 48,799 & 26.68 & 4.82 \\
 & Selected & 3,840 & 27.53 & 4.90 \\
\bottomrule
\end{tabular}
\end{table}

\FloatBarrier
\subsection{Twenty-seed sequence-baseline comparison}
\label{app:ehrlich-baselines}

This extends the main Ehrlich comparison to all 13 policies and full-budget
AUC, using the task and generator in Section~\ref{app:ehrlich-protocol}
and the settings in Table~\ref{tab:experimental-protocol}. Iterative arms
share the initial archive within each seed; unconditional generation starts
cold. Batch methods make 242 online acquisitions of eight, while Holo GA
uses four generations of 484.

Table~\ref{tab:ehrlich-headline} gives the budget-matched endpoint comparison.
\begin{table}[H]
\centering
\footnotesize
\setlength{\tabcolsep}{3pt}
\renewcommand{\arraystretch}{1.05}
\caption{\textbf{Ehrlich sequence optimisation.} Best reward after 2,000 charged
evaluations, mean $\pm$ sample SD over 20 seeds.
Bold marks the highest reported mean; $^*$ denotes a positive mean
difference and $p<0.05$ against every other method in that task's table.
Tests are exploratory and unadjusted (Supplementary Section~\ref{app:external-baseline-tests}). Paired tests are used.}
\label{tab:ehrlich-headline}
\begin{tabular}{@{}lr@{}}
\toprule
Method & Final best reward \\
\midrule
GenTuRBO & $\mathbf{0.1619\pm0.0719}^{*}$ \\
Hamming + GP/EI & $0.0563\pm0.0118$ \\
Holo GA & $0.0586\pm0.0100$ \\
AdaLead & $0.0555\pm0.0156$ \\
BO-EVO & $0.0336\pm0.0105$ \\
LaMBO-2 (online) & $0.0672\pm0.0114$ \\
\bottomrule
\end{tabular}
\par\smallskip\parbox{\columnwidth}{\footnotesize GenTuRBO, Hamming and AdaLead use the spaced-pair GP. AdaLead selects by predicted mean. Component comparisons: Supplementary Section~\ref{app:component-ehrlich}.}
\end{table}

Alongside unconditional and four conditional generator component arms, we
include an unconditional generator pool ranked by GP/EI, adaptive Hamming
mutation with random or GP/EI selection, and four sequence-search controls.
AdaLead uses the FLEXS-derived root-anchored protocol in
Section~\ref{app:protein-protocol}: threshold 0.05, no recombination,
128 predictions per round including roots, and eight paid candidates ranked
by predicted mean without a second EI pass. Here each site mutates with
probability $1/256$, drawing from all 20 tokens including its current value.
The shared initial archive and full-history spaced-pair GP are retained,
with refitting every eight updates. Invalid sequences consume budget and
receive zero-valued GP targets. The adapted
BO-EVO arm follows its
temperature-weighted parent sampling, non-empty per-site mutation, and
eight-child UCB screening policy (parent weights $\propto\exp(10f)$,
$\mu+0.2\sigma$ acquisition, 64 predictions for eight paid queries,
no crossover), but replaces the published warm start and GP
with the shared Ehrlich archive and sequence surrogate
\citep{hu2023protein}.  The matched-online LaMBO-2 arm uses reward-blind
Ehrlich generative pretraining, freshly learns its objective from the common
paid archive, and directly returns eight guided designs per round
\citep{gruver2023protein}. Pretraining samples 160,000 valid length-256
Markov sequences, uses 128,000 for training and 32,000 for validation, and
runs nine epochs at batch size 128. This matches the CTMC checkpoint's
294,912,000 training-token presentations, without exposing rewards or motifs.
The online settings are one initial epoch, one retraining epoch per round
on paid history, training batch size 128, one design step, two mutations per
step and at most eight guidance updates. These two rows are transparent
paid-oracle-budget-matched adaptations, not exact reproductions of their
published biological protocols.

Hamming proposals choose exactly $k$ distinct sites from all 256 positions
and replace each token uniformly with one of the 19 alternatives. The
rounded adaptive edit coordinate uses the same 2--128 bounds, initial 64,
and success/failure controller as GenTuRBO. It does not consult the hidden
Markov transition graph or repair invalid proposals. The GP arm shares
GenTuRBO's spaced-pair GP and selects eight of 128 proposals by marginal EI;
the random-selection arm uses the same proposal operator and pool size.

Holo GA uses the released DiscreteEvolution survival, recombination, and
mutation operators \citep{stanton2024closed}, initialised from the prepaid
shared archive.  Its population is 484, survival quantile is 0.01, and mutation
and recombination probabilities are both $1/256$.  Four generations use the
remaining 1,936 calls.  Its incumbent is updated only at the end of each
generation in trajectory and AUC summaries.

Two seed-7 conditional no-TR arms, using random and GP/EI selection, record a
cached-only proposal-collapse attempt with no acquisition or budget debit
before completing their 242 paid batches.  We retain the seed because each run
has exactly 1,936 online selections and 2,000 total paid evaluations.  Endpoint
and trajectory summaries use paid oracle calls; proposal-pool counts retain the
extra zero-acquisition attempt.

\begin{figure}[H]
    \centering
    \includegraphics[width=\textwidth]{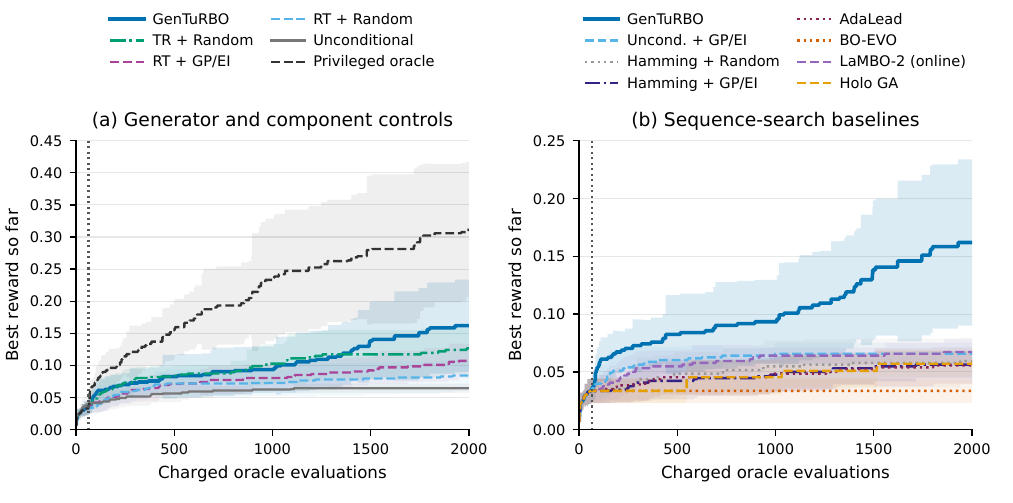}
    \caption{
    \textbf{Selected twenty-seed best-so-far trajectories.}
    Lines show the mean incumbent and bands show one sample standard deviation.
    The dotted line marks the end of the shared 64-evaluation initial archive
    for iterative policies; unconditional generation has no shared initial
    archive.  Panels use separate reward ranges.  The oracle-pool curve is
    privileged and incurs additional unbudgeted physical scores.  Holo GA is
    shown at completed-generation checkpoints. Hamming + GP/EI screens
    direct mutation proposals with the same spaced-pair GP/EI policy; AdaLead
    uses the shared GP with root-anchored rollouts and predicted-mean selection.
    }
    \label{fig:ehrlich-sequence-baselines-n20}
\end{figure}

Table~\ref{tab:ehrlich-sequence-baselines-n20} adds full-budget AUC and
supplementary controls to the main endpoints.
Figure~\ref{fig:ehrlich-sequence-baselines-n20} shows the corresponding
trajectories for generator controls and sequence-search baselines.
The paired GenTuRBO minus
RT + Random difference is $+0.0775$ (descriptive 95\% $t$ interval
$[0.0419,0.1131]$; 17 wins, two ties, one loss); against RT + GP it is
$+0.0549$ ($[0.0235,0.0862]$). These intervals are not adjusted for multiple
comparisons. Against Hamming + GP/EI, GenTuRBO wins 19 seeds and ties one.
Against AdaLead it wins all 20 seeds on both final reward and AUC.
The AdaLead cohort completed without exclusions in one packed
CPU-only allocation, using 1.291389 calculated node-hours including its
separate native smoke test; shared historical starts and training are excluded.

\begin{table}[H]
\centering
\small
\caption{
\textbf{Complete twenty-seed Ehrlich sequence-baseline comparison.}
Values are mean$\pm$sample standard deviation.  Full-budget AUC is the mean
incumbent over all 2,000 paid calls.  Applicable iterative arms share the same
64-evaluation initial archive within each seed; unconditional generation does
not.  The oracle-pool row is privileged and not matched in physical scoring
cost.  Holo GA improvements enter the AUC at completed-generation checkpoints.}
\label{tab:ehrlich-sequence-baselines-n20}
\begin{adjustbox}{max width=\textwidth}
\begin{tabular}{lrr}
\toprule
Policy & Final best & Full-budget AUC \\
\midrule
Unconditional + random & $0.0645\pm0.0071$ & $0.0590\pm0.0055$ \\
RT + Random & $0.0844\pm0.0245$ & $0.0710\pm0.0186$ \\
TR + Random & $0.1266\pm0.0368$ & $0.0974\pm0.0215$ \\
RT + GP & $0.1070\pm0.0270$ & $0.0791\pm0.0152$ \\
GenTuRBO (TR + GP) & $0.1619\pm0.0719$ & $0.1040\pm0.0358$ \\
Privileged oracle-pool & $0.3111\pm0.1053$ & $0.2115\pm0.0611$ \\
Unconditional pool + GP/EI & $0.0656\pm0.0096$ & $0.0610\pm0.0081$ \\
Hamming + random & $0.0594\pm0.0200$ & $0.0508\pm0.0166$ \\
Hamming + GP/EI & $0.0563\pm0.0118$ & $0.0470\pm0.0107$ \\
AdaLead & $0.0555\pm0.0156$ & $0.0470\pm0.0122$ \\
BO-EVO (adapted) & $0.0336\pm0.0105$ & $0.0334\pm0.0103$ \\
LaMBO-2 (matched-online) & $0.0672\pm0.0114$ & $0.0586\pm0.0069$ \\
Holo GA & $0.0586\pm0.0100$ & $0.0461\pm0.0095$ \\
\bottomrule
\end{tabular}
\end{adjustbox}
\end{table}

\FloatBarrier
\subsection{Frozen-archive kernel screening}
\label{app:ehrlich-static-kernels}

We next ask whether a sequence kernel that exposes local spaced motifs can
rank candidates more effectively than positional one-hot Tanimoto similarity.
For each condition, we construct 20 leave-one-source-seed-out splits with 512
training candidates and 256 held-out candidates, and evaluate top-eight
acquisition.  The six fixed kernels are positional one-hot; spaced pairs with
windows 32 and 48; a local spaced trigram with window 16; an equal-weight
pair-$W48$ plus trigram-$W16$ additive kernel; and a privileged task-aware
kernel using the true motif offsets.  In every GP, output scale is fixed to
one, kernel parameters are fixed, and only the constant mean and likelihood
noise are fitted.  All 240 fits completed without failure.

Table~\ref{tab:ehrlich-static-kernels} reports the frozen-archive results.
Reward scales differ across the two conditions, so raw regret and RMSE must not
be pooled across sequence lengths.  The structured kernels improve regret over
one-hot within both conditions, but no deployable kernel dominates every
metric.  The task-aware row is an intentionally privileged diagnostic and is
excluded from online use.  Spaced pair $W=32$ offers the clearest compromise:
it improves regret and rank correlation over one-hot, while avoiding the much
higher cost of the trigram and additive constructions.

\begin{table}[H]
\centering
\scriptsize
\caption{
\textbf{Frozen-archive Ehrlich kernel performance.}
Mean$\pm$sample standard deviation over 20 leave-one-source-seed-out splits
(512 train, 256 held out, top-eight acquisition).  Lower regret, RMSE, and NLPD
are better; higher recall and Spearman correlation are better.  The task-aware
kernel is privileged and non-deployable.}
\label{tab:ehrlich-static-kernels}
\begin{adjustbox}{max width=\textwidth}
\begin{tabular}{llrrrrr}
\toprule
Condition & Kernel & Regret & Recall & RMSE & Spearman & NLPD \\
\midrule
$L=128$ & Positional one-hot & $0.226\pm0.133$ & $0.013\pm0.056$ & $0.164\pm0.056$ & $-0.042\pm0.240$ & $-0.220\pm0.679$ \\
 & Spaced pair ($W=32$) & $0.129\pm0.112$ & $0.037\pm0.071$ & $0.138\pm0.052$ & $0.375\pm0.217$ & $-0.139\pm1.068$ \\
 & Spaced pair ($W=48$) & $0.118\pm0.114$ & $0.050\pm0.075$ & $0.139\pm0.053$ & $0.349\pm0.220$ & $-0.178\pm1.009$ \\
 & Spaced trigram ($W=16$) & $0.133\pm0.136$ & $0.044\pm0.061$ & $0.145\pm0.059$ & $0.268\pm0.172$ & $-0.368\pm0.735$ \\
 & Pair+trigram additive & $0.108\pm0.123$ & $0.056\pm0.086$ & $0.143\pm0.057$ & $0.326\pm0.189$ & $-0.327\pm0.837$ \\
 & Task-aware (privileged) & $0.049\pm0.073$ & $0.144\pm0.136$ & $0.132\pm0.050$ & $0.611\pm0.133$ & $-0.528\pm0.494$ \\
\midrule
$L=256$ & Positional one-hot & $0.021\pm0.020$ & $0.025\pm0.065$ & $0.022\pm0.009$ & $-0.022\pm0.182$ & $-2.260\pm0.518$ \\
 & Spaced pair ($W=32$) & $0.014\pm0.016$ & $0.031\pm0.090$ & $0.021\pm0.007$ & $0.069\pm0.193$ & $-1.658\pm1.336$ \\
 & Spaced pair ($W=48$) & $0.016\pm0.016$ & $0.037\pm0.100$ & $0.021\pm0.007$ & $0.056\pm0.212$ & $-1.696\pm1.261$ \\
 & Spaced trigram ($W=16$) & $0.019\pm0.015$ & $0.013\pm0.038$ & $0.022\pm0.008$ & $0.103\pm0.211$ & $-2.287\pm0.518$ \\
 & Pair+trigram additive & $0.015\pm0.016$ & $0.031\pm0.069$ & $0.022\pm0.008$ & $0.100\pm0.213$ & $-2.150\pm0.730$ \\
 & Task-aware (privileged) & $0.012\pm0.015$ & $0.013\pm0.038$ & $0.021\pm0.007$ & $-0.033\pm0.150$ & $-2.380\pm0.382$ \\
\bottomrule
\end{tabular}
\end{adjustbox}
\end{table}

\FloatBarrier
\subsection{Kernel fitting and inference cost}
\label{app:ehrlich-kernel-timing}

Table~\ref{tab:ehrlich-kernel-timing} measures CPU double-precision cost on
held-out source seed 0.  Fit time is measured once; the inference statistic is
the median of warm calls two and three after excluding the first of three
posterior calls.  Garbage collection occurs outside the timed intervals.  At
$L=256$, spaced pair $W=32$ costs 7.44 seconds to fit and 0.89 seconds for warm
inference, compared with 0.23 and 0.03 seconds for one-hot.  The additive
kernel is considerably more expensive (36.94 and 5.23 seconds) without a
commensurate frozen-archive advantage, supporting its exclusion from the
online comparison.

\begin{table}[H]
\centering
\small
\caption{
\textbf{Ehrlich kernel fitting and inference cost.}
CPU double-precision measurements on held-out source seed 0: one fit and the
median of warm posterior calls two and three, with the first call excluded and
garbage collection outside measured intervals.}
\label{tab:ehrlich-kernel-timing}
\begin{tabular}{l@{\hspace{2em}}lrr}
\toprule
Condition & Kernel & Fit (s) & Warm inference (s) \\
\midrule
$L=128$ & Positional one-hot & 0.344 & 0.012 \\
 & Spaced pair ($W=32$) & 3.315 & 0.395 \\
 & Spaced pair ($W=48$) & 5.049 & 0.573 \\
 & Spaced trigram ($W=16$) & 12.158 & 1.358 \\
 & Pair+trigram additive & 14.443 & 1.931 \\
 & Task-aware (privileged) & 0.595 & 0.046 \\
\midrule
$L=256$ & Positional one-hot & 0.229 & 0.027 \\
 & Spaced pair ($W=32$) & 7.436 & 0.892 \\
 & Spaced pair ($W=48$) & 11.879 & 1.387 \\
 & Spaced trigram ($W=16$) & 19.446 & 3.816 \\
 & Pair+trigram additive & 36.937 & 5.234 \\
 & Task-aware (privileged) & 3.296 & 0.321 \\
\bottomrule
\end{tabular}
\end{table}

\FloatBarrier
\subsection{Five-seed online confirmation}
\label{app:ehrlich-online-kernels}

\begin{figure}[H]
    \centering
    \includegraphics[width=\textwidth]{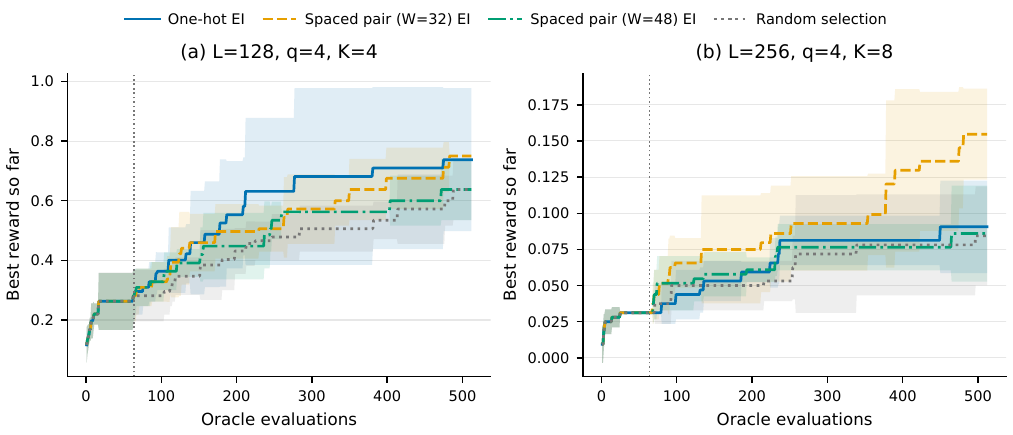}
    \caption{
    \textbf{Five-seed online Ehrlich kernel confirmation.}
    Curves show mean best reward so far and shaded bands show one sample
    standard deviation.  The dotted line marks the end of the shared
    64-evaluation initial archive.  Panels use separate, condition-specific
    reward scales.  The comparison is exploratory because it contains only
    five paired seeds.
    }
    \label{fig:ehrlich-online-kernels}
\end{figure}

Finally, we compare marginal expected improvement under positional one-hot and
the two spaced-pair kernels against random selection.  Within each of five
paired seeds, all arms share a byte-identical 64-evaluation initial archive.
Each arm then runs 56 batches with 128 generator proposals and eight oracle
evaluations, for 512 oracle calls in total.  The proposal policy uses the same
absolute edit-distance controller as the calibration study.  GP output scale
and all kernel parameters remain fixed; only the constant mean and likelihood
noise are fitted. Figure~\ref{fig:ehrlich-online-kernels} shows the
best-so-far trajectories, and Table~\ref{tab:ehrlich-online-kernels}
summarises endpoints, post-initial AUC and runtime.

At $L=256$, the observed mean paired $W=32$ minus one-hot difference is
$+0.0641$ for final best (four wins, one tie, no losses) and $+0.0270$ for
post-initial AUC.  At $L=128$, the observed final-best difference is $+0.0125$
(three wins and two losses), while the AUC difference is $-0.0503$.  The
$W=48$ kernel does not improve both conditions.  We therefore
retain $W=32$ as the only structured alternative for subsequent online study,
but treat this five-seed result as an exploratory confirmation rather than a
general kernel-performance claim.

\begin{table}[H]
\centering
\small
\caption{
\textbf{Matched online Ehrlich kernel comparison.}
Mean$\pm$sample standard deviation over five paired seeds.  Post-initial AUC is
the mean incumbent after the shared 64-evaluation archive and remains in
condition-specific reward units; scores are not pooled across sequence
lengths.}
\label{tab:ehrlich-online-kernels}
\begin{adjustbox}{max width=\textwidth}
\begin{tabular}{llrrr}
\toprule
Condition & Selection rule & Final best & Post-initial AUC & Runtime (s) \\
\midrule
$L=128$ & One-hot EI & $0.7375\pm0.2396$ & $0.6033\pm0.2124$ & 259.7 \\
 & Spaced pair ($W=32$) EI & $0.7500\pm0.0000$ & $0.5530\pm0.0586$ & 385.9 \\
 & Spaced pair ($W=48$) EI & $0.6375\pm0.1027$ & $0.5098\pm0.0376$ & 432.4 \\
 & Random selection & $0.6375\pm0.1027$ & $0.4691\pm0.0646$ & 254.2 \\
\midrule
$L=256$ & One-hot EI & $0.0906\pm0.0320$ & $0.0706\pm0.0092$ & 488.9 \\
 & Spaced pair ($W=32$) EI & $0.1547\pm0.0314$ & $0.0975\pm0.0241$ & 711.2 \\
 & Spaced pair ($W=48$) EI & $0.0859\pm0.0331$ & $0.0697\pm0.0110$ & 814.7 \\
 & Random selection & $0.0844\pm0.0342$ & $0.0646\pm0.0186$ & 481.0 \\
\bottomrule
\end{tabular}
\end{adjustbox}
\end{table}

\FloatBarrier

\FloatBarrier
\section{Additional Protein Diagnostics}
\label{app:protein-diagnostics}
\subsection{Current-cohort surrogate checks}
\label{app:protein-surrogate-checks}

For GFP-Medium, learned-noise Tanimoto gives $0.626\pm0.016$, while fixing
noise variance to 0.001 in standardised outcome units gives
$0.854\pm0.149$ across 20 seeds. The latter is the Tanimoto configuration
used in the main comparison. Section~\ref{app:protein-protocol} specifies
the cohort and hyperparameter-selection history. All 20 GFP-Hard Kermut GenTuRBO runs recorded
posterior-variance floors and zero EI in their first eight batches; all are
retained in the reported endpoints.

\FloatBarrier
\section{Sokoban Coordination Optimisation}
\label{app:sokoban}

\paragraph{Generator.}
The reward-blind Transformer follows \citet{baghal2026solvable}: six blocks,
width 256, eight attention heads, feedforward width 1,024 and 4.9M parameters.
The blocks use pre-normalisation, GELU and dropout 0.1, with learned row and
column embeddings and a sinusoidal-time MLP. The vocabulary has seven data
tiles (including box-on-goal and player-on-goal) plus an input-only mask.
It is trained from scratch on all 900,000 unfiltered Boxoban training boards
\citep{boxobanlevels}, holding out 100,000 validation boards, for 292,000
updates at effective batch size 1,536. Training seed 0 produces the single
checkpoint frozen for all 20 search seeds. Shuffled training epochs apply
uniformly sampled rotations and reflections of the square; validation is
unaugmented. AdamW uses peak learning rate $2.45\times10^{-4}$, weight decay
0.01, 125 updates of linear warmup followed by cosine decay to zero, and
global gradient-norm clipping at 1. Training uses FP16 autocast with gradient
scaling; cross-entropy is evaluated in float32. The 292,000 count denotes
successful optimiser updates, excluding overflow-skipped attempts. A fixed
4,096-board validation subset is evaluated every 1,000 updates; search uses
the final checkpoint rather than selecting weights by the switch objective.

For each board, draw $k\sim\mathrm{Unif}\{1,\ldots,100\}$ and independently
mask every cell with probability $k/100$. With $M$ the masked cells, the
per-board loss is
\[
\ell=\min(100/k,10)\frac{1}{|M|}\sum_{i\in M}
-\log p_\theta(x_i\mid x_{\rm masked},t=1-k/100).
\]
An empty mask contributes zero; losses are averaged across boards. Sampling
reveals one randomly selected masked cell per model call at temperature one;
unconditional generation begins with 100 masks. Conditional proposals choose
exactly the requested number of distinct cells uniformly from all 100 cells,
including borders, and preserve visible cells while filling the masks.
Diffusion receives no count correction, wall repair or solver-based rejection.
This is an independent training run using the reference architecture and
nominal update/batch budget, with the full unfiltered Boxoban split and the
augmentation and checkpoint conventions specified above.

\paragraph{Objective and protocol.}
For pushed-box identities $b_1,\ldots,b_P$, we optimise
\[
 B=\min_{\text{solutions}}\sum_{i=2}^{P}\mathbf{1}[b_i\ne b_{i-1}],
\]
the singleton-box decomposition measure of \citet{jarusek2010difficulty},
with the first box uncharged. Walking contributes zero, and consecutive
pushes of the same physical box remain in one group even if the player walks
between them. The minimum is over all solutions, rather than the switch count
of a push-optimal solution. The push-state solver tracks persistent box identities
and the last pushed box. Shared caps are 200,000 expanded
states, 400,000 discovered states and ten seconds per query. Invalid and
proven-unsolvable boards receive $-1$; censored (unresolved) scores are excluded from
GP labels, exact parents and the best-so-far score. All outcomes and queried
duplicates consume budget. Each arm is charged for the same 64 initial boards
within its seed, including failures; none are replaced. Canonical identity
quotients grid symmetries and freely reachable player positions. A bounded
objective with a valid witness can count as solvable without an exact score.

\paragraph{Operators.}
GenTuRBO uses the best exactly scored solvable board as its parent, generates
128 proposals and selects eight by marginal EI, without local penalisation.
Its CPU-double one-hot Tanimoto GP retains all finite paid observations, refits mean/noise
every eight updates, and fixes outputscale to one with jitter $10^{-4}$.
Masking starts at 20 cells, expands by 1.5 after an improving batch and
contracts by 0.75 otherwise, within 5--40 cells. Failure at the lower bound
resets it to 20.

AdaLead uses the shared FLEXS-derived protocol in
Section~\ref{app:protein-protocol}: threshold 0.05, eight measured-order roots,
fixed-root rollouts, 128 predictions including roots, and eight paid queries
ranked by predicted mean. Each of the 100 cells mutates independently with
probability $1/100$, drawing from all seven public tiles including its current
value. There is no recombination, inventory repair or feasibility screening.
Initial parent multiplicity is preserved. The full-history Tanimoto GP and
refitting schedule match GenTuRBO. AdaLead's 128 model predictions are
distinct from GenTuRBO's 128 generated proposals.

Hamming + GP/EI shares GenTuRBO's parent, pool/batch sizes, surrogate,
refitting schedule and adaptive controller. It uniformly selects exactly
$r$ distinct cells from all 100 positions and replaces each with one of the
six non-current public tiles; the mask token is excluded. There is no inventory
or border repair or feasibility prescreening. Distance is measured before
canonicalisation; generator mask counts need not equal realised edits.

Direct-grid GA has population 32, tournament size three, elitist survival,
crossover probability 0.5 and two grid mutations. It evaluates eight offspring
before each population update, without a surrogate. Rectangle crossover,
tile swaps and wall/floor flips operate over all 100 cells, including borders.
Crossover normalises the public inventory over the whole grid but never forces
border walls or tests solvability; all invalid offspring are charged. These
direct-grid arms use no diffusion after initialisation. Independent sampling
evaluates every raw generated board.

\paragraph{Results and validation.}
All 100 runs completed their 2,000-query budgets.
Table~\ref{tab:sokoban-headline} reports the endpoint comparison, while
Table~\ref{tab:sokoban-results} adds maximum scores, solvability and unresolved
query counts. Solvability covers all charged queries, including starts and
duplicates. The reported endpoint is each seed's highest exactly scored
$B$ at 2,000 queries, summarised by the mean and sample SD across seeds.
GenTuRBO wins all 20 pairs against Hamming and GA, and 19 with one tie
against AdaLead. Against independent sampling it wins 15, ties three and
loses two. Hamming never improves on its best initial board.
All 38,720 post-initial Hamming queries fail: 38,714 are
structurally invalid and six are proven unsolvable. No seeds are excluded.
All 1,000 displayed-method top-ten solution certificates replayed successfully.
Independent uniform-cost search without A* heuristics or deadlock pruning
confirms the displayed GenTuRBO winner's $B=18$ after 1,748,750 expansions.
It has a verified
30-push, switch-optimal witness; push optimality is not claimed.

\begin{table}[H]
\centering
\footnotesize
\setlength{\tabcolsep}{3pt}
\renewcommand{\arraystretch}{1.05}
\caption{\textbf{Sokoban coordination optimisation.} Best minimum box-switch count after 2,000 charged
evaluations, mean $\pm$ sample SD over 20 seeds.
Bold marks the highest reported mean; $^*$ denotes a positive mean
difference and $p<0.05$ against every other method in that task's table.
Tests are exploratory and unadjusted (Supplementary Section~\ref{app:external-baseline-tests}). Paired tests are used.}
\label{tab:sokoban-headline}
\begin{tabular}{@{}lr@{}}
\toprule
Method & Best $B$ \\
\midrule
GenTuRBO & $\mathbf{12.450\pm2.438}^{*}$ \\
Hamming + GP/EI & $5.350\pm1.137$ \\
Direct-grid GA & $7.500\pm1.701$ \\
AdaLead & $6.950\pm1.791$ \\
\bottomrule
\end{tabular}
\par\smallskip\parbox{\columnwidth}{\footnotesize GenTuRBO and Hamming share the best-parent, marginal-EI policy and full-history Tanimoto GP. AdaLead selects root-anchored rollouts by predicted mean. Direct-grid GA evaluates eight offspring per update. All operators can edit border cells (Supplementary Section~\ref{app:sokoban}).}
\end{table}

\begin{table}[H]
\centering
\begin{tabular}{lrrrr}
\toprule
Method & Final best $B$ & Maximum $B$ & Solvable & Unresolved / 40,000 \\
\midrule
Independent sampling & $9.450\pm1.669$ & 13 & 88.09\% & 146 \\
Direct-grid GA & $7.500\pm1.701$ & 11 & 11.22\% & 62 \\
AdaLead & $6.950\pm1.791$ & 11 & 19.79\% & 4 \\
Hamming + GP/EI & $5.350\pm1.137$ & 9 & 2.85\% & 4 \\
GenTuRBO & $\mathbf{12.450\pm2.438}$ & $\mathbf{18}$ & $\mathbf{88.76\%}$ & 127 \\
\bottomrule
\end{tabular}
\caption{\textbf{Sokoban outcomes and solver coverage.} Final best $B$ is
each seed's best exact score after 2,000 queries, reported as mean $\pm$ sample
SD over 20 matched seeds. Maximum $B$ is the highest exact score across all
seeds; solvability and unresolved counts cover all 40,000
charged queries per method. Bold marks the highest score or solvability;
unresolved counts describe solver coverage rather than an optimisation target.}
\label{tab:sokoban-results}
\end{table}

\paragraph{Compute and limits.}
Search uses CPU float32 generator inference and CPU-double GPs. Calculated
node-hours are 0.997431 for 20 GenTuRBO seeds and 0.027500 for AdaLead.
GA and Hamming shared a 365-second allocation costing 0.025347 node-hours,
including smoke tests. Totals exclude shared initial generation, training and
unrelated probes; packed steps are counted only through their allocation.
These estimates are not portal-reconciled, and differing workloads preclude
an intrinsic runtime comparison. Human difficulty, training-set novelty and
separate effects of adaptive masking and acquisition remain unestablished.

\FloatBarrier
\section{Comparisons with External Baselines}
\label{app:external-baseline-tests}
We compare each task's highest-mean reported method with every other method
in its reported results table, including both GenTuRBO protein kernels and
Hamming + GP/EI. All comparisons use 20 runs per method at 2,000 charged
evaluations. The $t$-tests are paired for Ehrlich and Sokoban, which share
seed-specific initial archives, and Welch for proteins, whose seeds share
one fixed task-level archive rather than seed-specific experimental blocks.
An asterisk in a results table requires a positive mean difference and
two-sided $p<0.05$ against \emph{every} other method reported in that column.
Results-table asterisks and pointwise confidence intervals are unadjusted.
Leading methods are selected from observed means, so these comparisons
are exploratory and are not selection-adjusted confirmatory tests.

As a protein-only robustness check, Holm correction is applied across all
84 pairwise comparisons among the seven methods on the four protein tasks,
not just the contrasts involving the observed leading methods. GFP-Medium
Hamming + GP/EI remains significantly above each of the six other methods
under this correction (largest adjusted $p=2.40\times10^{-7}$).
The results-table asterisks retain the unadjusted exploratory convention above;
the correction does not adjust for earlier hyperparameter or protocol selection.

\begin{table}[H]
\centering
\footnotesize
\setlength{\tabcolsep}{3pt}
\renewcommand{\arraystretch}{1.05}
\caption{\textbf{Comparisons for the highest-mean reported method in each task.}
$\Delta$ is the leading method minus the comparator; intervals are pointwise
95\% CIs and $p$-values are unadjusted and two-sided. Paired tests are used
for Ehrlich and Sokoban, Welch tests for proteins. All reported methods,
including both GenTuRBO protein kernels and Hamming + GP/EI, receive the same
comparison rule. The leading method is selected from observed means, so
these tests are exploratory. GFP-Medium GenTuRBO (Tanimoto) and Hamming + GP/EI
both use fixed noise variance 0.001 in standardised outcome units.}
\label{tab:all-external-tests}
\begin{adjustbox}{max width=\textwidth}
\begin{tabular}{@{}lllrrr@{}}
\toprule
Task & Leading method & Comparator & $\Delta$ & 95\% CI & $p$ \\
\midrule
Ehrlich & GenTuRBO & Hamming + GP/EI & $+0.1057$ & $[0.0713, 0.1400]$ & $3.54\times10^{-6}$ \\
Ehrlich & GenTuRBO & Holo GA & $+0.1033$ & $[0.0713, 0.1354]$ & $1.89\times10^{-6}$ \\
Ehrlich & GenTuRBO & AdaLead & $+0.1064$ & $[0.0746, 0.1383]$ & $1.16\times10^{-6}$ \\
Ehrlich & GenTuRBO & BO-EVO & $+0.1283$ & $[0.0930, 0.1636]$ & $3.53\times10^{-7}$ \\
Ehrlich & GenTuRBO & LaMBO-2 (online) & $+0.0947$ & $[0.0619, 0.1276]$ & $8.41\times10^{-6}$ \\
\midrule
GFP-M & Hamming + GP/EI & GenTuRBO (Kermut) & $+0.1782$ & $[0.1332, 0.2232]$ & $1.15\times10^{-9}$ \\
GFP-M & Hamming + GP/EI & GenTuRBO (Tanimoto) & $+0.3074$ & $[0.2325, 0.3824]$ & $6.00\times10^{-9}$ \\
GFP-M & Hamming + GP/EI & Holo GA & $+0.2970$ & $[0.2618, 0.3321]$ & $2.94\times10^{-17}$ \\
GFP-M & Hamming + GP/EI & AdaLead & $+0.2552$ & $[0.2183, 0.2921]$ & $1.04\times10^{-15}$ \\
GFP-M & Hamming + GP/EI & BO-EVO & $+0.5375$ & $[0.5062, 0.5688]$ & $3.23\times10^{-19}$ \\
GFP-M & Hamming + GP/EI & LaMBO-2 (online) & $+0.3865$ & $[0.3519, 0.4210]$ & $4.20\times10^{-20}$ \\
\midrule
GFP-H & Hamming + GP/EI & GenTuRBO (Kermut) & $+0.1685$ & $[0.1282, 0.2089]$ & $3.82\times10^{-10}$ \\
GFP-H & Hamming + GP/EI & GenTuRBO (Tanimoto) & $+0.0813$ & $[0.0364, 0.1262]$ & $8.03\times10^{-4}$ \\
GFP-H & Hamming + GP/EI & Holo GA & $+0.5809$ & $[0.5531, 0.6086]$ & $2.82\times10^{-24}$ \\
GFP-H & Hamming + GP/EI & AdaLead & $+0.2309$ & $[0.1921, 0.2698]$ & $1.96\times10^{-14}$ \\
GFP-H & Hamming + GP/EI & BO-EVO & $+0.6972$ & $[0.6596, 0.7348]$ & $1.58\times10^{-31}$ \\
GFP-H & Hamming + GP/EI & LaMBO-2 (online) & $+0.3292$ & $[0.2952, 0.3633]$ & $2.99\times10^{-21}$ \\
\midrule
AAV-M & GenTuRBO (Tanimoto) & GenTuRBO (Kermut) & $+0.0268$ & $[0.0052, 0.0483]$ & $0.0162$ \\
AAV-M & GenTuRBO (Tanimoto) & Hamming + GP/EI & $+0.0081$ & $[-0.0163, 0.0325]$ & $0.5053$ \\
AAV-M & GenTuRBO (Tanimoto) & Holo GA & $+0.1768$ & $[0.1568, 0.1968]$ & $1.46\times10^{-19}$ \\
AAV-M & GenTuRBO (Tanimoto) & AdaLead & $+0.0308$ & $[0.0020, 0.0595]$ & $0.0368$ \\
AAV-M & GenTuRBO (Tanimoto) & BO-EVO & $+0.2256$ & $[0.2008, 0.2503]$ & $3.99\times10^{-20}$ \\
AAV-M & GenTuRBO (Tanimoto) & LaMBO-2 (online) & $+0.1006$ & $[0.0788, 0.1224]$ & $2.17\times10^{-11}$ \\
\midrule
AAV-H & Hamming + GP/EI & GenTuRBO (Kermut) & $+0.0394$ & $[-0.0105, 0.0892]$ & $0.1177$ \\
AAV-H & Hamming + GP/EI & GenTuRBO (Tanimoto) & $+0.0297$ & $[-0.0277, 0.0871]$ & $0.2988$ \\
AAV-H & Hamming + GP/EI & Holo GA & $+0.1429$ & $[0.1112, 0.1746]$ & $9.15\times10^{-10}$ \\
AAV-H & Hamming + GP/EI & AdaLead & $+0.0741$ & $[0.0429, 0.1053]$ & $5.06\times10^{-5}$ \\
AAV-H & Hamming + GP/EI & BO-EVO & $+0.2217$ & $[0.1859, 0.2575]$ & $1.35\times10^{-14}$ \\
AAV-H & Hamming + GP/EI & LaMBO-2 (online) & $+0.1000$ & $[0.0635, 0.1364]$ & $2.64\times10^{-6}$ \\
\midrule
Sokoban & GenTuRBO & Hamming + GP/EI & $+7.100$ & $[6.027, 8.173]$ & $2.21\times10^{-11}$ \\
Sokoban & GenTuRBO & Direct-grid GA & $+4.950$ & $[3.648, 6.252]$ & $1.81\times10^{-7}$ \\
Sokoban & GenTuRBO & AdaLead & $+5.500$ & $[4.146, 6.854]$ & $6.70\times10^{-8}$ \\
\bottomrule
\end{tabular}
\end{adjustbox}
\end{table}

\FloatBarrier
\section{Component Analyses and Diagnostic Controls}
\label{app:component-analyses}

This section mirrors the main benchmark comparison and separates the
component matrices from the external-baseline endpoints. Every condition
below uses 20 seeds and 2,000 charged oracle evaluations, including
initialisation; entries are mean $\pm$ sample standard deviation.
The generator is frozen. RT samples the noising time uniformly, TR adapts
the corruption scale, Random selects proposals uniformly, and GP uses
expected improvement. GenTuRBO combines TR and GP selection. These are
descriptive component comparisons; the external-baseline tests are in
Section~\ref{app:external-baseline-tests}.

\subsection{Ehrlich sequence optimisation}
\label{app:component-ehrlich}

The four conditional-generator arms form a complete comparison of noising
control and proposal selection, sharing the task, generator, initial
archive and proposal budget. Table~\ref{tab:ehrlich-components} shows that
adaptive noising increases mean final reward with either selection policy,
and GP selection increases the mean with either noising policy.

\begin{table}[H]
\centering
\caption{\textbf{Ehrlich component comparison.} Final best reward under
the shared 2,000-evaluation protocol. All GP arms use the spaced-pair kernel.}
\label{tab:ehrlich-components}
\begin{tabular}{llr}
\toprule
Noising scale & Proposal selection & Final best reward \\
\midrule
Random time & Random & $0.0844\pm0.0245$ \\
Random time & GP & $0.1070\pm0.0270$ \\
Adaptive & Random & $0.1266\pm0.0368$ \\
Adaptive & GP (GenTuRBO) & $0.1619\pm0.0719$ \\
\bottomrule
\end{tabular}
\end{table}

With GP selection held fixed, adaptive noising improves mean final reward
from 0.1070 to 0.1619 (15 paired wins, four ties, one loss). With random
selection, it improves the mean from 0.0844 to 0.1266 (13 wins, six ties,
one loss). Within adaptive proposal pools, GP selection increases the mean
from 0.1266 to 0.1619, with eight wins, eight ties and four losses. Thus the
component pattern is complementary in the means but is not uniform across
seeds. Section~\ref{app:ehrlich-baselines} reports all 13 policies,
full-budget AUC and paired contrasts, while
Section~\ref{app:ehrlich-locality} reports corruption-to-edit calibration.

\subsection{Protein fitness optimisation}
\label{app:component-protein}

Table~\ref{tab:protein-components} retains all current 20-seed component
and surrogate conditions from the combined endpoint table. The four
Kermut/Random arms compare noising control and proposal selection; the
additional Tanimoto arms compare surrogate configurations with GP selection.
Protein Random and GP arms use different proposal-pool sizes
(Table~\ref{tab:experimental-protocol}), so their contrast includes both
pool size and selection policy.

\begin{table}[H]
\centering
\small
\caption{\textbf{Protein component and surrogate comparison.} Best
normalised CNN proxy score at 2,000 charged evaluations. GP denotes Kermut
unless Tanimoto is named; Random arms do not use a surrogate for selection.
$^\dagger$Both GFP-Medium Tanimoto arms fix noise variance to 0.001 in
standardised outcome units; other Tanimoto tasks learn noise.}
\label{tab:protein-components}
\begin{tabular}{@{}lcccc@{}}
\toprule
Method & GFP-Medium & GFP-Hard & AAV-Medium & AAV-Hard \\
\midrule
RT + Random & $0.623\pm0.004$ & $0.300\pm0.050$ & $0.565\pm0.058$ & $0.445\pm0.043$ \\
RT + GP & $0.627\pm0.019$ & $0.284\pm0.083$ & $0.759\pm0.049$ & $0.668\pm0.061$ \\
TR + Random & $0.672\pm0.057$ & $0.644\pm0.077$ & $0.680\pm0.040$ & $0.534\pm0.061$ \\
GenTuRBO (Kermut) & $0.983\pm0.074$ & $0.845\pm0.069$ & $0.806\pm0.033$ & $0.757\pm0.090$ \\
RT + GP (Tanimoto)$^\dagger$ & $0.623\pm0.004$ & $0.368\pm0.064$ & $0.780\pm0.048$ & $0.697\pm0.090$ \\
GenTuRBO (Tanimoto)$^\dagger$ & $0.854\pm0.149$ & $0.933\pm0.081$ & $0.832\pm0.035$ & $0.766\pm0.109$ \\
\bottomrule
\end{tabular}
\end{table}

With Kermut held fixed, adaptive noising increases the mean from 0.627 to
0.983 on GFP-Medium, 0.284 to 0.845 on GFP-Hard, 0.759 to 0.806 on
AAV-Medium, and 0.668 to 0.757 on AAV-Hard. GP selection also improves the
mean over random selection within adaptive pools on all four tasks. This
does not hold for every component contrast: under random-time noising,
Kermut GP selection is slightly worse than random selection on GFP-Hard
(0.284 versus 0.300).

Kermut has the highest GenTuRBO mean on GFP-Medium; Tanimoto has the highest
means on GFP-Hard and both AAV tasks. The mean gain from adaptive noising
is also positive for every Tanimoto task. These comparisons retain the
GFP-Medium fixed-noise configuration and all runs with variance floors;
Section~\ref{app:protein-surrogate-checks} describes the learned-noise
comparison and variance-floor diagnostics. Kernel comparisons do not by
themselves isolate the effect of acquisition from the trajectories it induces.

The protein results also highlight an important limitation of proxy-based
optimisation benchmarks: high predicted fitness does not establish that
proposed sequences remain experimentally viable, because biological validity
is not measured independently of the learned objective. By contrast, Ehrlich
and Sokoban allow structural validity to be evaluated directly. On these
tasks, generator-defined locality provides a clear advantage over direct
Hamming proposals, suggesting that the generative prior is particularly
valuable when important structural constraints are not captured explicitly
by the optimisation objective.

\subsection{Sokoban coordination optimisation}
\label{app:component-sokoban}

The comparisons in Table~\ref{tab:sokoban-results} support the value of local
optimisation under the reported protocols. GenTuRBO improves over independent
generator sampling in 15 paired seeds, with three ties and two losses. AdaLead uses the same
full-history Tanimoto GP and paid batch size, but its uniform mutations
frequently produce invalid boards; GenTuRBO wins 19 pairs and ties one.
Of the 38,720 post-initial charged queries per method, AdaLead spends
29,058 on structurally invalid boards (75.05\%); whole-grid GA spends
27,530 (71.10\%).

Hamming + GP/EI uses the same best parent, 128-proposal pool, marginal-EI
batch of eight, full-history GP and adaptive controller as GenTuRBO. Replacing
only generator proposals with exact-distance uniform mutations over all
100 cells finds no new solvable boards, so its solvability and best scores
arise entirely from shared initial boards. Whole-grid GA evaluates eight
offspring per update and is also below GenTuRBO in all 20 paired seeds. These controls
do not form a full noising-by-selection component matrix or rule out more
structured hand-designed operators. Section~\ref{app:sokoban} reports the
five-method comparison, solver coverage, canonicalisation and validation
of the current GenTuRBO winner with $B=18$.

\FloatBarrier
\section{Comparison with Continuous TuRBO}
\label{app:turbo-genturbo}
\begin{table}[H]
\centering
\small
\setlength{\tabcolsep}{3pt}
\renewcommand{\arraystretch}{1.12}
\caption{\textbf{From TuRBO to GenTuRBO.} Comparison with a single
TuRBO region.}
\label{tab:turbo-genturbo}
\begin{tabular}{@{}p{0.24\columnwidth}p{0.30\columnwidth}p{0.40\columnwidth}@{}}
\toprule
\raggedright\textbf{Component} & \raggedright\textbf{TuRBO} & \raggedright\textbf{GenTuRBO} \tabularnewline
\midrule
\raggedright Centre & \raggedright Incumbent & \raggedright Incumbent or archive parents \tabularnewline
\raggedright Local region & \raggedright Geometric trust region & \raggedright Generator-defined proposal distribution \tabularnewline
\raggedright Scale & \raggedright Region size & \raggedright Noising depth \tabularnewline
\raggedright Adaptation & \raggedright Success/failure & \raggedright Success/failure \tabularnewline
\raggedright Surrogate & \raggedright GP & \raggedright GP \tabularnewline
\raggedright Acquisition & \raggedright Candidates within region & \raggedright Generated candidate pool \tabularnewline
\raggedright Proposal prior & \raggedright No pretrained proposal model & \raggedright Fixed pretrained generator \tabularnewline
\bottomrule
\end{tabular}
\end{table}

\end{document}